\documentclass[letterpaper]{article} 
\usepackage{aaai24}  
\usepackage{times}  
\usepackage{helvet}  
\usepackage{courier}  
\usepackage[hyphens]{url}  
\usepackage{graphicx} 
\usepackage{natbib}  
\usepackage{caption} 
\usepackage{algorithm}
\usepackage{algorithmic}

\usepackage{multirow}
\usepackage{booktabs}
\usepackage{colortbl}  
\usepackage{xcolor}
\usepackage{array}   
\usepackage{utfsym}
\usepackage{makecell}
\usepackage{bm}
\usepackage{amsfonts}
\usepackage{amsmath}
\usepackage{mathtools}

\usepackage{newfloat}
\usepackage{listings}
\DeclareCaptionStyle{ruled}{labelfont=normalfont,labelsep=colon,strut=off} 
\floatstyle{ruled}
\newfloat{listing}{tb}{lst}{}
\floatname{listing}{Listing}
\title{EVE: Efficient Vision-Language Pre-training with Masked Prediction and Modality-Aware MoE}
\author {
    Junyi Chen\textsuperscript{\rm 1}\thanks{This work was done while Junyi was an intern at ByteDance.},
    Longteng Guo\textsuperscript{\rm 2},
    Jia Sun\textsuperscript{\rm 3},
    Shuai Shao\textsuperscript{\rm 3},
    Zehuan Yuan\textsuperscript{\rm 3},
    Liang Lin\textsuperscript{\rm 1},
    Dongyu Zhang\textsuperscript{\rm 1}\thanks{Corresponding author.}
}
\affiliations {
    \textsuperscript{\rm 1}Sun Yat-sen University\\
    \textsuperscript{\rm 2}Institute of Automation, Chinese Academy of Sciences (CASIA)\\
    \textsuperscript{\rm 3}Bytedance Inc\\
    chenjy765@mail2.sysu.edu.cn, longteng.guo@nlpr.ia.ac.cn, \{sunjia.ly, shaoshuai.0516, yuanzehuan\}@bytedance.com\\linliang@ieee.org, zhangdy27@mail.sysu.edu.cn
}

\usepackage{bibentry}

\begin{document}

\maketitle

\begin{abstract}
Building scalable vision-language models to learn from diverse, multimodal data remains an open challenge. In this paper, we introduce an Efficient Vision-languagE foundation model, namely EVE, which is one unified multimodal Transformer pre-trained solely by one unified pre-training task. Specifically, EVE encodes both vision and language within a shared Transformer network integrated with modality-aware sparse Mixture-of-Experts (MoE) modules, which capture modality-specific information by selectively switching to different experts. To unify pre-training tasks of vision and language, EVE performs masked signal modeling on image-text pairs to reconstruct masked signals, i.e., image pixels and text tokens, given visible signals. This simple yet effective pre-training objective accelerates training by 3.5x compared to the model pre-trained with Image-Text Contrastive and Image-Text Matching losses. Owing to the combination of the unified architecture and pre-training task, EVE is easy to scale up, enabling better downstream performance with fewer resources and faster training speed. Despite its simplicity, EVE achieves state-of-the-art performance on various vision-language downstream tasks, including visual question answering, visual reasoning, and image-text retrieval.
\end{abstract}

\section{Introduction}
Vision-Language Pre-training aims to learn a general multimodal representation that can be transferred to various vision-language downstream tasks, such as vision-language understanding and image-text retrieval.
A vision-language foundation model should have excellent performance while being easy to train and scale up, which can be achieved through the model architecture and the pre-training tasks.

\begin{figure}
    \centering
    \includegraphics[width=\linewidth]{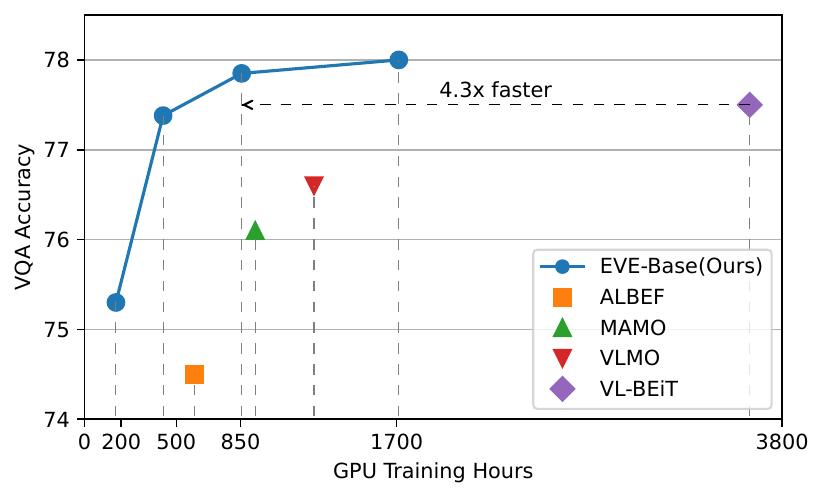}
    \caption{Performance of different models on VQA test-dev under different training hours. Training hours of all models are reproduced by us on A100 GPUs.}
    \label{fig:gputime}
\end{figure}

The model architectures of recent methods can be roughly divided into two categories: dual-encoder architecture and unified architecture.
Dual-encoder methods~\cite{clip,xvlm} employ modality-specific models (e.g. BERT~\cite{bert}, ViT~\cite{vit}) to encode different modalities separately and a fusion module to integrate them.
As for the fusion module, some methods~\cite{clip} employ shallow fusion (e.g., dot product) for the interaction of vision and language.
Some alternative methods~\cite{xvlm} use deep neural networks, such as Transformer Encoders, to perform deep fusion on modality interaction, but lead to difficulties in scaling up and low efficiency.
Unified methods~\cite{vilt, simvlm} use a modality-shared Transformer to encode different modalities jointly. This approach simplifies the framework and improves the speed, helping with model scaling up. However, they overlook the inherent gap between modalities, leading to lower overall performance.
Image is continuous, redundant, and low-level on the raw signals, while text is discrete, refined, and high-level. 
Directly using a shared Transformer to encode different modalities with semantic gap poses problems.
Therefore, it is necessary to consider the differences between different modalities carefully.

Previous methods also have explored numerous pre-training tasks for vision-language pre-training, including Image-Text Contrastive Learning~\cite{clip}, Image-Text Matching~\cite{albef}, Word-Patch Alignment~\cite{uniter},  Masked Language Modeling~\cite{vlbert}, Masked Image Modeling~\cite{vlbeit}, and so on. They have been widely used to improve vision-language pre-training.
While incorporating more pre-training tasks can enhance performance, adding too many tasks can also lead to some problems.
Foremost, it significantly prolongs the pre-training time and increases the computational resources required.
Additionally, it necessitates manual weight adjustments for different objectives.
Furthermore, excessive pre-training objectives can result in a reduction in the model's scalability, which is crucial in designing pre-training models, as the recent success has shown in large language models~\cite{instructgpt, emerge, chatbridge}. Therefore, it is necessary to use effective and scalable pre-training tasks.

In this paper, we propose an Efficient Vision-languagE foundation model (EVE) with a unified modality-aware Transformer pre-trained with a single unified pretraining task, i.e., masked signal modeling.

In terms of model architecture, we use a unified modality-aware Transformer and revisit the integration of Mixture-of-Experts in vision-language pre-training. 
We employ a shared Multi-Head Self-Attention module and a Modality-Aware MoE module for the modality-aware Transformer to encode and fuse various modalities jointly.
Using a unified shared Transformer is more concise and flexible, which simplifies the extension to additional modalities and facilitates cross-modal alignment.
By incorporating MoE, we can take into account the differences between modalities and capture more modality-specific information.
We also introduce a modality routing technique in MoE that enables the router select more appropriate experts for processing.

In terms of pre-training tasks, we propose a unified masked signal modeling technique combining masked pixel and language modeling, which significantly improves training speed and reduces scaling difficulty. Some methods~\cite{beit3, maskvlm, mamo} have applied generative pre-training paradigm to vision-language pre-training. While they either add the generative objective with other complex objectives like ITC and ITM~\cite{maskvlm} or employ more complicated targets such as visual tokens~\cite{beit3} or momentum features~\cite{mamo}, which require a nontrivial visual tokenizer or momentum model. All of these increase the complexity of pre-training. In contrast to them, we just utilize the \emph{raw signals} from the image-text pairs themselves to minimize the complexity of pre-training and achieve better scalability.
Pre-training speed is 3.5x faster than incorporating ITC and ITM.

EVE can greatly enhance pre-training speed, as shown in Figure~\ref{fig:gputime}. It decreases the demand for extensive computational resources while being easy to scale up.
We demonstrate the effectiveness of EVE on various vision-language downstream tasks, including visual question answering, visual reasoning, and image-text retrieval. EVE achieves state-of-the-art performance on Image-Text Retrieval and Vision-Language Understanding (VQA and NLVR2) tasks.

Our contributions are summarized as follows:
\begin{itemize}
    \item We introduce EVE, an efficient vision-language foundation model that achieves state-of-the-art performance while improving training speed, with one unified multimodal Transformer and one unified pre-training task.
    \item We integrate Modality-Aware MoE with a shared multimodal Transformer to achieve a more profound fusion of different modalities and capture more modality-specific information simultaneously, resulting in better performance and faster inference speed within a unified architecture.
    \item We propose a unified masked signal modeling technique, simplifying vision-language pre-training into a single unified objective, resulting in significantly improved pre-training speed and competitive performance.
\end{itemize}

\section{Related Work}
Model architecture and pre-training tasks are crucial factors in the representation learning of vision-language.

\paragraph{Model Architecture}
Dual-encoder with a fusion module~\cite{albef, opt, fiber, mamo} performs well on vision-language tasks but with higher time and architecture complexity.
Unified architecture methods~\cite{vilt,simvlm,vlmo,vlbeit} can flexibly encode different modalities as a fusion encoder or process a single modality as a unimodal encoder, demonstrating faster inference speed and promising performance.
Some of them~\cite{vilt, simvlm} use a shared standard Transformer~\cite{trans} to jointly encode different modalities, while they ignore the modality gap and lead to worse performance.
Others~\cite{vlmo,vlbeit} use MoME Transformer instead and prove that shared attention is better for multimodal learning. However, MoME Transformer uses modality-shared FFN in the deep layers may neglect some modality-specific information.

Considering the simplicity, effectiveness, and flexibility of the unified architecture, we adopt a unified architecture with Modality-Aware MoE to better capture modality specifics during fusion for multimodal representation learning.
We achieve state-of-the-art performance with approximately the same inference cost.
\begin{figure*}
    \centering
    \includegraphics[width=0.98\linewidth]{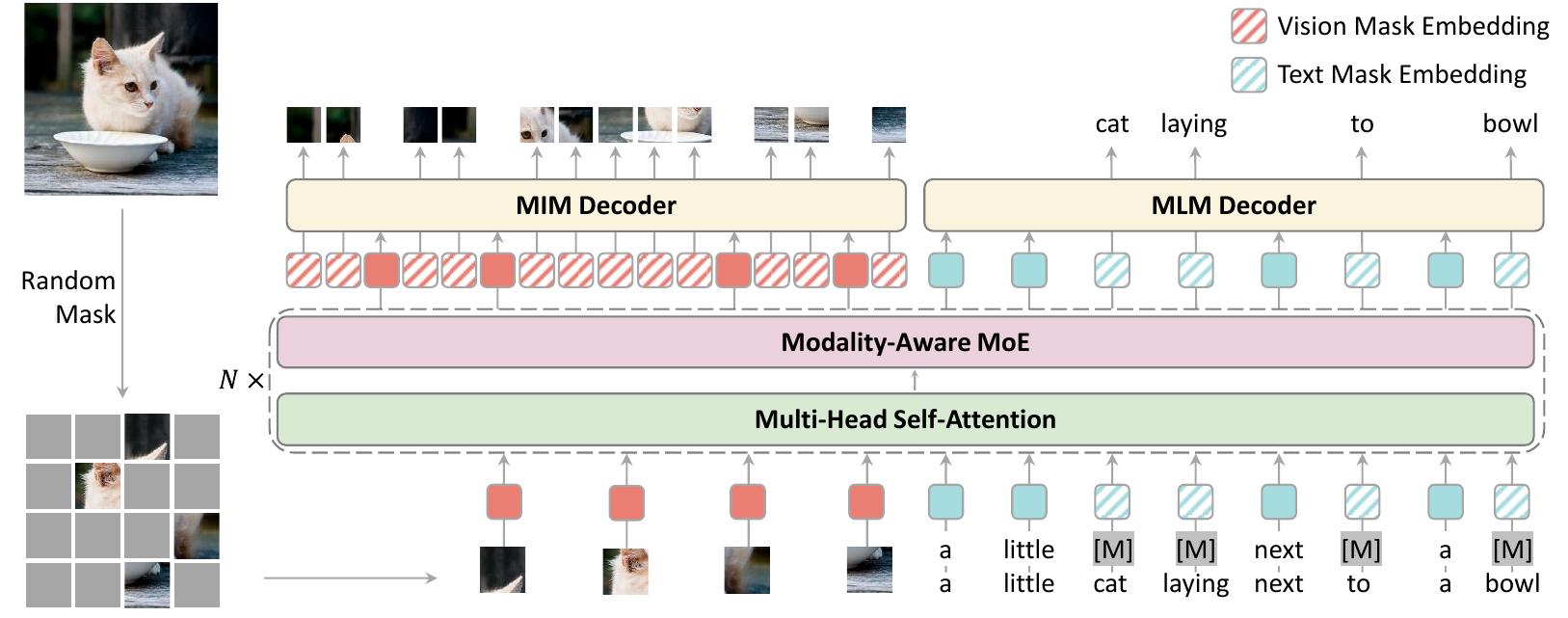}
    \caption{Overview of EVE and Masked Signal Modeling. We use a unified architecture with shared attention and Modality-Aware MoE for EVE and a single unified masked signal modeling for pre-training. We employ random masking on both image and text. Masked image and complete text are used in masked image modeling, vice versa.}
    \label{fig:pretrain}
\end{figure*}

\paragraph{Masked Signal Modeling}
Recently, several methods~\cite{vlbeit,mamo,vlmae,DAVINCI,m3ae} explore the "mask then predict" paradigm in the vision for vision-language pre-training.
While VLBEiT~\cite{vlbeit} introduces training on the visual modality through masked image modeling, their reconstruction target is the visual token, which may significantly influence performance depending on the visual tokenizer employed.
DAVINCI~\cite{DAVINCI} extends prefix language modeling further to vision, but it also uses the discrete visual token as the target.
MAMO~\cite{mamo} enriches multimodal representation by using momentum features in masked representation modeling, which relies heavily on a momentum teacher model to avoid divergence.
Some methods~\cite{maskvlm, vlmae, vlc} use masked pixel modeling, but they all require additional costly pre-training tasks such as ITC~\cite{clip} and ITM~\cite{visualbert}. Among these methods, VLMAE~\cite{vlmae} only applies masked pixel modeling to the image encoder.
M3AE~\cite{m3ae} leverages a unified Image-Language masking approach to mask and reconstruct both images and text simultaneously, but it is not used in multimodal downstream tasks.

We unify masked pixel and language modeling into masked signal modeling, reconstructing masked raw signals from visible signals.
This simplifies and accelerates training, achieving better performance and scalability.

\paragraph{Mixture-of-Experts (MoE)}
Mixture-of-Experts has been extensively explored in computer vision~\cite{vmoe} and natural language processing~\cite{auxloss, gshard}.
These methods generally aim to improve performance by learning a better routing using auxiliary losses~\cite{gshard, zloss}, converting it into a linear assignment problem~\cite{base}, or making it differentiable~\cite{diff}.
MoE seems well-suited for multimodal learning, but the differences between modalities present some challenges.
LIMoE~\cite{LIMoE} involves more auxiliary losses to balance different modalities, uni-perceiver-moe~\cite{uni-moe} employs conditional MoE, VLMO~\cite{vlmo} and VL-MoE~\cite{vlmoe} use shared expert in the deep layers.

However, existing methods increase complexity or limit performance due to manual routing and ignoring modality information.
Therefore, we propose Modality-Aware MoE as a simple way to apply MoE to multimodal learning. We simplify the auxiliary loss and capture more modality specifics by expert switching.

\section{Methods}
\subsection{Backbone Network}
As shown in Figure~\ref{fig:pretrain}, we adopt a unified multimodal Transformer with shared attention and Modality-Aware Mixture-of-Experts as the backbone network, which is capable of encoding different modalities.
After pre-training, the model can be utilized as either a fusion encoder or a unimodal encoder for various downstream tasks through fine-tuning. 

For Image $\boldsymbol{I}$, following VIT~\cite{vit}, we first split the Image $\boldsymbol{I}$ into $N$ patches with a patch size of $P$. The resulting {\small$N=HW/P^2$} patches
are projected into a shared embedding space using a linear projector. A special token $\boldsymbol{I}_{\text{cls}}$ is added at the beginning of all visual tokens. We employ learnable visual position embeddings $\boldsymbol{I}_{\text{pos}}$ and visual type embeddings $\boldsymbol{I}_{\text{type}}$ on visual tokens.
Image embedding can be summarized as follows.
\begin{equation}
\fontsize{9.5}{9.5}\selectfont
\boldsymbol{I}_{\text{emb}} = [\boldsymbol{I}_{\text{cls}}, \boldsymbol{I}_1, \dots, \boldsymbol{I}_N] + \boldsymbol{I}_{\text{pos}} + \boldsymbol{I}_{\text{type}}
\end{equation}

For Text $T$, following BERT~\cite{bert}, we tokenize the text into discrete tokens with the maximum length of $n$ and project them into the joint embedding space. We add a special token $\boldsymbol{T}_{\text{cls}}$ at the beginning of all text tokens and use learnable text position embeddings $\boldsymbol{T}_{\text{pos}}$ and text type embeddings $\boldsymbol{T}_{\text{type}}$ for text encoding
Text embedding can be summarized as follows.
\begin{equation}
\fontsize{9.5}{9.5}\selectfont
\boldsymbol{T}_{\text{emb}} = [\boldsymbol{T}_{\text{cls}}, \boldsymbol{T}_1, \dots, \boldsymbol{T}_n] + \boldsymbol{T}_{\text{pos}} + \boldsymbol{T}_{\text{type}}
\end{equation}

We concatenate $\boldsymbol{I}_{\text{emb}}$ and $\boldsymbol{T}_{\text{emb}}$ as the input to the model:
\begin{equation}
\fontsize{9.5}{9.5}\selectfont
\mathbb{P}_{\text{emb}} = [\boldsymbol{I}_{\text{emb}}, \boldsymbol{T}_{\text{emb}}]
\end{equation}

\subsection{Modality-Aware Mixture-of-Experts}
\begin{figure}
    \centering
    \includegraphics[width=\linewidth]{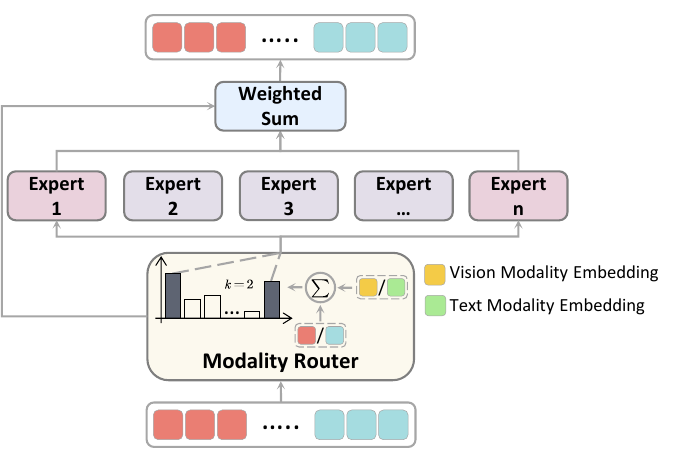}
    \caption{Architecture of Modality-Aware MoE.}
    \label{fig:model_moe}
\end{figure}

Multimodal learning differs significantly from unimodal learning, as the differences between modalities cannot be ignored.
Using the same Feed-Forward Network for all modalities can lead to inappropriate fusion of modalities, resulting in degraded performance.
Conversely, using modality-specific MoE in all layers may not benefit the alignment of different modalities.
Therefore, we propose the Modality-Aware Mixture-of-Experts (MoE) as shown in Figure~\ref{fig:model_moe}, which incorporates the modality routing technique on top of the general MoE to capture modality-specific information while fusing by selectively switching to different experts.

In the general MoE, each MoE block typically consists of $N$ experts, and each input token is processed by $k$ experts selected from the $N$ experts.
A lightweight router $g$ is used to select the $k$ experts for each token, which employs a simple linear-softmax predictor to calculate the routing weight.
This can be formulated as:
\begin{equation}
\fontsize{9.5}{9.5}\selectfont
g(\mathbf{x})=\text{softmax}\left(\mathbf{W} \cdot \mathbf{x}\right)
\end{equation}
$\mathbf{W}\in \mathbb{R}^{D \times N}$ is a learnable projector for input $\mathbf{x}\in \mathbb{R}^{D}$.

The final output of the MoE block is the weighted average of the $k$ selected experts, which can be formulated as:
\begin{equation}
\fontsize{9.5}{9.5}\selectfont
\text{MoE}(\mathbf{x})=\sum_{i=1}^{k} g(\mathbf{x})_{i} \cdot \text{FFN}_{i}(\mathbf{i})
\end{equation}

\subsubsection{Modality Routing}
General MoE does not impose any restrictions on the router, which can easily lead to unbalanced routing. LIMoE~\cite{LIMoE} points out that this phenomenon can be exacerbated in multimodal learning due to the difference in token count across different modalities.

To address this issue, we propose a modality-aware routing approach to enhance the router.
We adopt a best-effort strategy for routing to preserve all tokens while explicitly providing modality information to the router by adding modality-specific embeddings.
The new routing function can be formulated as follows:
\begin{equation}
\fontsize{9.5}{9.5}\selectfont
g(\mathbf{x})=\text{softmax}\left(\mathbf{W} \cdot (\mathbf{x}+\mathbf{b}_{m})\right)
\end{equation}

Here, we use modality-specific embeddings $\mathbf{b}_{m} \in \mathbb{R}^{D}$ for different modalities, i.e., $\mathbf{b}_{I}$ for images and $\mathbf{b}_{T}$ for text.

\subsubsection{Auxiliary Loss}
In addition to modality routing, we use a single simple auxiliary loss to balance routing and avoid carefully tuning the weight. Following~\citet{auxloss}, we add Load-Balancing Loss as the auxiliary loss to train the router. It can be formulated as follows:
\begin{equation}
\fontsize{9.5}{9.5}\selectfont
\mathcal{L}_{aux}= \alpha\cdot N\sum_{i}^{N} f_i \times p_i
\end{equation}

This objective encourages uniform routing of tokens, where $N$ denotes the number of experts, $f_i$ denotes the fraction of tokens dispatched to the {\small$i^{\text{th}}$} expert, and $p_i$ denotes the average routing weight for the {\small$i^{\text{th}}$} expert.
The weight $\alpha$ is a hyperparameter that we set at 0.001 by default to avoid overwhelming other objectives.
 
Considering efficiency, we use a soft router with top-$k=2$ in the deep layers and a hard router in the shallow layers.
An MoE module equipped with a hard router has the same number of experts as the number of modalities. The hard router directly selects the corresponding expert based on the modality of each token.

\subsection{Pre-training Task: Masked Signal Modeling}

Previous multimodal models~\cite{albef, clip, vlmo, visualbert, mamo} typically involve complex pre-training tasks like Image-Text Contrastive Learning (ITC)~\cite{clip}, Image-Text Matching (ITM)~\cite{visualbert}, and Masked Representation Modeling (MRM)~\cite{mamo}.
These methods have shown good performance, but pre-training still requires significant computational resources, and is challenging to scale up.

Table~\ref{tab:time} shows the efficiency comparison between different pre-training tasks, which indicates a significant difference in time consumption and batch size.
Compared to pre-training without ITC and ITM, including them requires four times more computational resources to achieve a similar speed.
Moreover, ITC and ITM tasks are similar to other contrastive learning-based methods that typically require a larger batch size to achieve better performance.
Incorporating additional pre-training tasks can significantly decrease training speed, increase training difficulty, and have an impact on the scalability of the model.

\begin{table}
\centering
\fontsize{9}{9}\selectfont
\begin{tabular}{ccccc|c|c}
\toprule
\multicolumn{5}{c|}{Pre-training Tasks} & \multirow{3}{*}{\makecell[c]{Batch\\size}} & \multirow{3}{*}{\makecell[c]{Time}} \\
MLM & ITC & ITM & \makecell[c]{MIM\\Token} & \makecell[c]{MIM\\Pixel} &  &  \\ \midrule
\usym{2713} & &  & &   \usym{2713} & \textbf{224} & \textbf{2.14h} \\ 
\usym{2713} &  &  & \usym{2713} &  & 152 & 3.09h \\ 
\usym{2713} & \usym{2713}  &  &  & \usym{2713}  & 132               & 3.26h \\ 
\usym{2713} & \usym{2713} &  \usym{2713} &  &    & 80                & 6.88h \\ 
\usym{2713} & \usym{2713} & \usym{2713} &  & \usym{2713}  & 64                & 7.73h \\ 
\bottomrule 
\end{tabular}
\caption{Maximum batch size per GPU and pre-training time per epoch of different pre-training tasks on 8 A100 GPUs with the same architecture as EVE-Base. We add vision mask tokens in the encoder during masked token modeling.}
\label{tab:time}
\end{table}

Thus, we pre-train our model with only one unified masked signal modeling objective on image-text pairs to reconstruct masked signals by visible signals as shown in Figure~\ref{fig:pretrain}.
Specifically, masked signal modeling combines masked image modeling and masked language modeling, and only utilizes the raw signals from image-text pairs themselves without relying on any additional techniques.
We use masked image and complete text in masked image modeling, while complete image and masked text in masked language modeling.
Despite its simplicity, our approach achieves competitive performance compared to previous methods and can be easily scaled up.

In this section, we use $h(\cdot)$ and $\theta(\cdot)$ to denote the encoder and the decoder. 
$\hat{I}$ and $\hat{T}$ are represented for masked image and masked text. $D$ indicates the dataset.

\subsubsection{Masked Language Modeling (MLM)}

Following BERT~\cite{bert}, we randomly mask some of the text tokens and predict them based on the information provided by the image and corrupted text. The Masked Language Modeling (MLM) objective can be formulated as follows:
\begin{equation}
\mathcal{L}_{mlm}=\mathbb{E}_{(I, T) \sim D} \ell_{mlm}\left(\theta_t\left(h(I, \hat{T})\right), T\right)
\end{equation}

$\ell_{mlm}$ computes the cross-entropy loss between the prediction probability $P_{mlm}$, obtained from the text decoder $g_t$, and the ground truth on each masked token. We use a two-layer MLP with a softmax layer as the text decoder.

\subsubsection{Masked Image Modeling (MIM)}
Previous methods~\cite{mamo, xfm, beit3} have typically employed semantically rich visual features obtained from the model itself or discrete visual tokens obtained from visual tokenizers as the targets for MIM.
However, both approaches have their drawbacks.
Training visual tokenizers~\cite{dalle, beitv2} is a challenging task as different tokenizers can have varying impacts on performance and may lead to error propagation.
Meanwhile, using visual features~\cite{mamo, xfm} requires either applying momentum distillation techniques or employing other loss functions and techniques to prevent the model from diverging during training.
These MIM targets make the overall framework more complex. 

In visual self-supervised learning, some works use other information as the MIM targets, such as RGB pixels~\cite{mae}, scene depth~\cite{multimae}, HOG~\cite{maskfeat}, etc.
However, using targets such as scene depth and HOG requires additional techniques, which increases the complexity of the training process.
In order to maintain simplicity and effectiveness, we choose to utilize the image pixels themselves as the reconstruction target.

Following MAE~\cite{mae}, we adopt an asymmetric design for MIM, where only observed image patches and all text tokens are fed into the encoder. A lightweight decoder is used to reconstruct raw pixels on masked positions from partial image representation and masked tokens, as shown in Figure \ref{fig:pretrain}.
We use multiple Transformer blocks with narrower hidden widths as the decoder. The MIM objective can be formulated as:
\begin{equation}
\fontsize{9.5}{9.5}\selectfont
\mathcal{L}_{mim}=\mathbb{E}_{(I, T) \sim D} \ell_{mim}\left(\theta_i\left(h(\hat{I}, T)\right), I\right)
\end{equation}

$\ell_{mim}$ calculates the mean square error between the raw pixels and the reconstructed result generated by the image decoder. We compute the loss on masked image patches.

The overall objective of masked signal modeling is:
\begin{equation}
\fontsize{9.5}{9.5}\selectfont
\mathcal{L} = \mathcal{L}_{mlm} + \mathcal{L}_{mim}
\end{equation}

\section{Experiments}
\subsection{Pre-training Datasets}
Following Previous methods, we pre-train EVE on four widely used public datasets:  MSCOCO Captions~\cite{mscoco}, Visual Genome~\cite{vg}, SBU Captions~\cite{sbu} and Conceptual Captions~\cite{cc}. There are about 4M images and 10M image-text pairs in all datasets. Since some downstream tasks are based on COCO, we exclude all images in the test sets of downstream tasks from the pre-training data. We also pre-train EVE-Large on a larger dataset with 21M image-text pairs by adding CC12M~\cite{cc12m}.

\subsection{Implementation Details}
EVE-Base has 12 Transformer blocks and EVE-Large has 24 Transformer blocks. We employ a soft router with 32 experts in EVE-Base on top-2 blocks, EVE-Large on top-3 blocks, and a hard router on the other blocks. We pre-train EVE-Base for 480k steps with a batch size of 2048 and EVE-Large with the same batch size for 280k steps. We use AdamW~\cite{adamw} optimizer. The peak learning rate is 5e-4 for EVE-Base and 2e-4 for EVE-Large. During pre-training, the image resolution is $224\times 224$. We use random resized cropping and  horizontal flipping for data augmentation. We mask 75\% of image in MIM and 50\% of text in MLM. EVE is initialized with BEiTv2. More details are provided in Appendix.

\begin{table*}
\centering
\fontsize{9.3}{9.3}\selectfont
\begin{tabular}{lr|cc|cc|cc|cc}
\toprule
\multicolumn{1}{l}{\multirow{2}{*}{Model}} & \multicolumn{1}{c|}{\multirow{2}{*}{\#Images}} & \multicolumn{2}{c|}{VQA} & \multicolumn{2}{c|}{NLVR2} & \multicolumn{2}{c|}{COCO} & \multicolumn{2}{c}{Flickr30K} \\
\multicolumn{1}{c}{}    & \multicolumn{1}{c|}{}  & test-dev & test-std & dev  & test-P & TR@1 & IR@1 & TR@1 & IR@1 \\ 
\midrule
ALBEF \cite{albef}    & 4M & 74.54    & 74.70      & 80.24   & 80.50 & 73.1 &56.8 & 94.3 & 82.8     \\
Triple \cite{triple}    & 4M & 74.90 & 74.92 & 80.54 & 81.33  & 75.6 & 59.0 & 94.9 & 84.0     \\
Codebook \cite{codebook} & 4M   & 74.86 & 74.97 & 80.50 & 80.84 & 75.3 & 58.7 & 95.1 & 83.3      \\
METER \cite{meter}    & 4M   & 77.68 & 77.64 & 82.33 & 83.05  & 76.2 & 57.1 & 94.3 & 82.2    \\
MAMO \cite{mamo}      & 4M & 76.12       & 76.20      & 81.86       & 81.53 & 77.1 & 60.3 & 95.6 & \textcolor{red}{\textbf{85.4}}      \\
VLMO \cite{vlmo}      & 4M &  76.64       & 76.89      & 82.77       & 83.34 & 74.8 & 57.2 & 92.3 & 79.3     \\
VL-BEiT \cite{vlbeit} & 4M &  77.53       & 77.75      & 81.93       & 82.66 & 79.5 & 61.5 & \textcolor{red}{\textbf{95.8}} & 83.9     \\
VLMAE \cite{vlmae}    & 4M &  75.30        & 75.40       & 80.50        & 81.20  & 77.3 & 59.6 & 95.2 & 83.6    \\
MaskVLM \cite{maskvlm} & 4M &  75.45       & 75.40      & 81.58      & 81.98 & 76.3 & 60.1 & 95.6 & 84.5     \\
VLC-Base \cite{albef} & 5.6M & 74.02 & 74.00 & 77.70 & 79.04  & 72.4 & 50.7 & 89.2 & 71.3\\
\rowcolor{gray!40} DAVINCI \cite{DAVINCI} & 631.8M &   76.32 & 76.44      &  80.03 & 80.25  & - & - & - & -     \\ 
\rowcolor{gray!40} SimVLM-Base \cite{simvlm}   & 1.8B &  77.87 & 78.14 & 81.72 & 81.77  & - & - & - & -   \\ 
\rowcolor{gray!40} BEiT3-Base \cite{beit3}    & 3.1B &  77.65       & -          & 83.60       & 84.40   & 79.1 & 61.4 & 96.3 & 86.2    \\ 
EVE-Base (Ours)       & 4M &  \textcolor{red}{\textbf{78.00}}  & \textcolor{red}{\textbf{78.02}}   & \textcolor{red}{\textbf{83.34}} & \textcolor{red}{\textbf{83.93}} & \textcolor{red}{\textbf{79.6}} & \textcolor{red}{\textbf{62.0}} & 95.6 & 84.1  \\
\bottomrule
\end{tabular}
\caption{Comparison with state-of-the-art base-size models on VQA, NLVR2, MSCOCO and Flickr30K. \colorbox{gray!40}{Gray} lines indicate the model pre-trained with much more data (more than 400M).}
\label{tab:base}
\end{table*}

\subsection{Vision-Language Downstream Tasks}
We evaluate our pre-trained model on three common Vision-Language Tasks. More implementation details and comparison on inference speed are provided in Appendix.

\subsubsection{\textbf{Visual Question Answering (VQA)}} VQA requires the model to predict an answer based on the given image and question. We use VQA2.0 dataset~\cite{vqa} to evaluate our model.
Following previous work~\cite{vlmo}, we view the task as a classification task.

\subsubsection{\textbf{Natural Language for Visual Reasoning (NLVR2)}} Given a sentence and two images, NLVR2 asks the model to judge whether the sentence accurately describes the relationship between the two images. We evaluate our model on NLVR2 dataset~\cite{nlvr}.
Following~\citet{uniter}, we convert the triplet input into two image-text pairs with the same text description and different images.

\subsubsection{\textbf{Image-Text Retrieval}} Retrieval task contains two sub-tasks: Image-to-Text Retrieval (TR) and Text-to-Image Retrieval (IR). We evaluate the model on widely used Flickr30K~\cite{flickr} and MSCOCO~\cite{mscoco} benchmarks following Karpathy split~\cite{Karpathy}.
Following~\citet{albef}, we apply ITC and ITM losses in the fine-tuning stage and we use rerank strategy during inference.

\begin{table}
\centering
\fontsize{9.5}{9.5}\selectfont
\begin{tabular}{l|cc|cc|c} 
\toprule
\multirow{2}{*}{MIM Target} & \multicolumn{2}{c|}{NLVR2} & \multicolumn{2}{c|}{Flickr30K} & \multirow{2}{*}{VQA}  \\
                            & dev & test-P                                             & TR & IR                        \\ 
\midrule
BEiTv2 Token  & 78.0 & 78.5 & 92.6 & 78.3 & 76.6 \\
DALL-E Token  & $\times$ & $\times$ & 92.4 & 77.4 & 75.8 \\
Pixel (Ours)   & \textbf{79.7} & \textbf{80.1} & \textbf{93.9} & \textbf{80.7} & \textbf{77.3} \\
\bottomrule
\end{tabular}
\caption{Ablation study on MIM target. $\times$ denotes divergence during fine-tuning.} 
\label{tab:mimtarget}
\end{table}

\subsection{Results on Downstream Tasks}
We present the results of VQA, NLVR2, COCO, and Flickr30K with state-of-the-art base models in Table \ref{tab:base} and large models in Table \ref{tab:large}. We report the accuracy for VQA and NLVR2, top-1 recall for TR and IR.
\paragraph{\textbf{Results on Vision-Language Understanding}}
EVE-Base outperforms all previous methods on Understanding tasks and even marginally outperforms BEiT3-Base \cite{beit3} pre-trained with 3.1B data on VQA. EVE-Base outperforms VLMO \cite{vlmo}, which also employs a unified architecture with more pre-training objectives by 1.77\% on VQA test-dev and 0.70\% on NLVR2 test-P.
EVE-Large4M shows similar performance to SimVLM-Large \cite{simvlm}, whereas EVE-Large16M surpasses SimVLM-Huge which is larger and pre-trained on much more data.

\paragraph{\textbf{Results on Image-Text Retrieval}}
EVE-Base achieves competitive results on Flickr and outperforms the previous state-of-the-art methods on COCO. Compared to VLMO, EVE-Base achieves improvements of 6.42\% on COCO text retrieval R@1 and 8.39\% on COCO image retrieval R@1.
In addition, EVE-Large demonstrates better performance on both COCO and Flickr30K to other Large or even Huge models with very limited data. 
Notably, Image-Text Contrastive Learning and Image-Text Matching are not involved in the pre-training of EVE.

\begin{table*}
\centering
\fontsize{9.3}{9.3}\selectfont
\begin{tabular}{lr|cc|cc|cc|cc}
\toprule
\multicolumn{1}{l}{\multirow{2}{*}{Model}} & \multicolumn{1}{c|}{\multirow{2}{*}{\#Images}} & \multicolumn{2}{c|}{VQA} & \multicolumn{2}{c|}{NLVR2} & \multicolumn{2}{c|}{COCO} & \multicolumn{2}{c}{Flickr30K} \\
\multicolumn{1}{c}{}    & \multicolumn{1}{c|}{}  & test-dev & test-std & dev  & test-P & TR@1 & IR@1 & TR@1 & IR@1 \\ 
\midrule
VinVL-Large \cite{vinvl} & 8.9M & 76.52 & 76.60 & 82.67 & 83.98 & 75.4 & 58.8 & - & - \\
BLIP-CapFiltL \cite{blip} & 129M & 78.25 & 78.32 & 82.15 & 82.24 & 81.2 & 64.1 & 97.2 & 87.5 \\
BLIP-Large \cite{blip} & 129M & - & - & - & - & 82.4 & 65.1 & 97.4 & 87.6 \\
Uni-PerceiverMoE-L \cite{uni-moe} & 44.1M & - & - & - & - & 74.7 & 58.3 & 94.1 & 83.7 \\ 
FILIP-Large \cite{filip}  & 340M & - & - & - & - & 78.9 & 61.2 & 96.6 & 87.1 \\
Prismer-Large \cite{prismer}  & 12.7M & 78.4 & 78.5 & - & - & - & - & - & - \\
\rowcolor{gray!40} GIT \cite{git}  & 800M & 75.5 & - & - & - & - & - & - & - \\
\rowcolor{gray!40} ALIGN-Large \cite{align}  & 1.8B & - & - & - & - & 77.0 & 59.9 & 95.3 & 84.9 \\
\rowcolor{gray!40} SimVLM-Large \cite{simvlm} & 1.8B & 79.32 & 79.56 & 84.13 & 84.84 & - & - & - & - \\
\rowcolor{gray!40} SimVLM-Huge \cite{simvlm} & 1.8B &  80.03 & 80.34 & 84.53 & 85.15 & - & - & - & -\\
\rowcolor{gray!40} Florence-Huge \cite{florence} &  900M &  80.16 & 80.36 & - & - & 81.8 & 63.2 & 97.2 & 87.9 \\
EVE-Large (Ours) & 4M &  79.25 & 79.20 & 84.03 & 84.69 & 82.5 & 65.2 & 96.3 & 86.3  \\
EVE-Large (Ours) & 16M & \textcolor{red}{\textbf{80.17}} & \textcolor{red}{\textbf{80.18}} & \textcolor{red}{\textbf{85.63}} & \textcolor{red}{\textbf{86.22}} & \textcolor{red}{\textbf{83.5}} & \textcolor{red}{\textbf{66.7}} & \textcolor{red}{\textbf{98.0}} & \textcolor{red}{\textbf{87.9}}  \\
\bottomrule
\end{tabular}
\caption{Comparison with state-of-the-art large-size models on VQA, NLVR2, MSCOCO and Flickr30K. \colorbox{gray!40}{Gray} lines indicate the model pre-trained with much more data (more than 400M).}
\label{tab:large}
\end{table*}

\subsection{Ablation Studies}
For all ablation studies, we pre-train the model for 25 epochs with a similar architecture to EVE-Base and report accuracy on NLVR2, VQA dev set, and top-1 recall on Flickr30K. We use the soft router with top-$k=2$ by default. We present some more ablation studies in Appendix.

\begin{figure}
  \centering
  \begin{minipage}{0.48\linewidth}
    \includegraphics[width=0.98\textwidth]{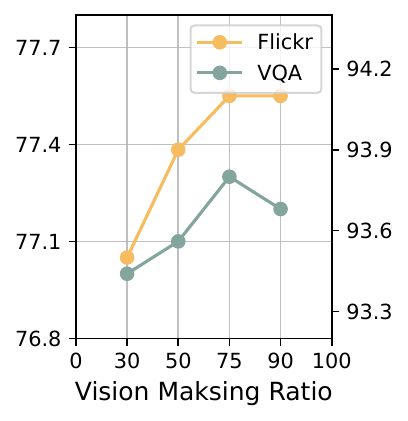}
    \label{fig:visionmask}
  \end{minipage}
  \begin{minipage}{0.48\linewidth}
    \includegraphics[width=0.98\textwidth]{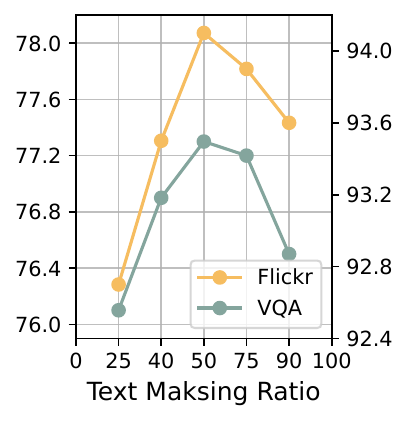}
    \label{fig:textmask}
  \end{minipage}
  \caption{Ablation study on masking ratio. Left and right y-axis denote VQA accuracy and Flickr mean recall.}
  \label{fig:maskratio}
\end{figure}

\subsubsection{\textbf{MIM Target}} We compare different MIM targets in Table~\ref{tab:mimtarget}, including image token and pixel. We use the tokenizer from BEiT v2~\cite{beitv2} and DALL-E~\cite{dalle}. It is observed that reconstructing pixels is better than reconstructing image tokens in all tasks. Using a more complex MIM target does not achieve the expected effect.

\paragraph{\textbf{Masking Ratio}}
In Figure~\ref{fig:maskratio}, we investigate the impact of different masking ratios on both vision and language.
Results indicate that a higher vision masking ratio leads to improved performance.
We hypothesize that the raw signals are highly redundant for image, and a higher masking ratio is needed to facilitate representation learning.
The noteworthy difference from previous work~\cite{mamo} is that we achieve better performance at a higher text masking ratio.
Our interpretation is that with a more profound integration of vision and language, the model can more easily predict masked text tokens with the aid of vision.

\begin{figure}
  \centering
  \begin{minipage}{0.49\linewidth}
    \includegraphics[width=0.98\textwidth]{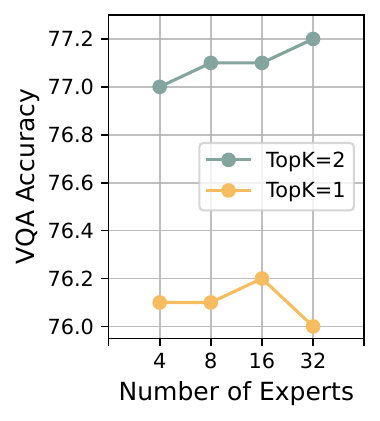}
    \label{fig:moe_vqa}
  \end{minipage}
  \begin{minipage}{0.49\linewidth}
    \includegraphics[width=0.98\textwidth]{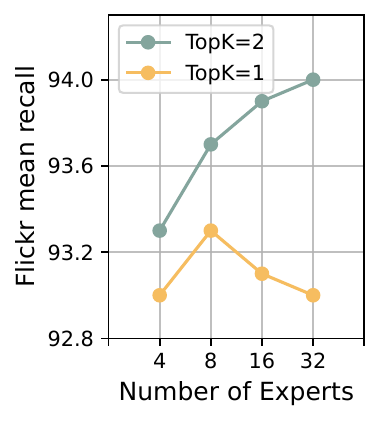}
    \label{fig:moe_flickr}
  \end{minipage}
  \caption{Ablation study on the number of experts and top-$k$ design. We use soft router in [8, 10, 12] Transformer blocks.}
  \label{fig:moeexpers}
\end{figure}

\paragraph{\textbf{Number of Experts and Top-K}}
The number of experts and the selection of top-$k$ are crucial aspects of MoE design, as they determine the model's parameters, computational complexity, and performance.
Figure~\ref{fig:moeexpers} clearly demonstrates that performance deteriorates as the number of selected experts decreases from 2 to 1. When $k=1$, increasing the number of experts can actually lead to a decrease in performance, which is more evident in retrieval tasks.
When $k=2$, increasing the number of experts leads to corresponding improvements in the performance of both VQA and retrieval tasks, with a more significant improvement observed in the retrieval task.

\begin{table}
\centering
\fontsize{9}{9}\selectfont
\begin{tabular}{cc|cc|cc|c}
\toprule
\multicolumn{2}{c|}{Tasks} & \multicolumn{2}{c|}{NLVR2} & \multicolumn{2}{c|}{Flickr30K} & \multirow{2}{*}{VQA} \\
MIM & MLM & dev & test-P & TR & IR &  \\ \midrule
\usym{2713} &  & 57.2  & 57.4  & 30.4 & 22.9 & 60.9  \\
 & \usym{2713} & 78.8  & 79.3  & 92.2 & 79.2 & 77.0 \\
\usym{2713}$^\dagger$ & \usym{2713}$^\dagger$  & 75.4 & 75.7 & 88.6 & 74.2 & 74.6 \\ 
\usym{2713} & \usym{2713} & \textbf{79.7}  & \textbf{80.1} & \textbf{93.9} & \textbf{80.7} & \textbf{77.3}  \\
\bottomrule
\end{tabular}
\caption{Ablation study on MIM and MLM. $\dagger$ denotes the model is pre-trained by MIM and MLM simultaneously with masked image and text inputs. Masking ratio is set to 50\% for both image and text in $\dagger$, but 75\% for image in others.}
\label{tab:pre-training}
\end{table}

\begin{table}
\centering
\fontsize{9}{9}\selectfont
\begin{tabular}{cccc|cc|c}
\toprule
\multicolumn{4}{c|}{Pre-training Tasks} & \multicolumn{2}{c|}{Flickr30K} & \multirow{2}{*}{VQA} \\
MIM & MLM & ITC & ITM & TR & IR &   \\ \midrule
\usym{2713} & \usym{2713} & \usym{2713} & & 94.0 & 80.0 & 76.8 \\
\usym{2713} & \usym{2713} &  & \usym{2713} & 94.0 & 80.7 & 77.0 \\
\usym{2713} & \usym{2713} & \usym{2713} & \usym{2713} & 94.2 & 80.8 & 77.1 \\
\usym{2713} & \usym{2713} & & & \textbf{94.4} & \textbf{81.2} & \textbf{77.4} \\
\bottomrule 
\end{tabular}
\caption{Ablation study on more pre-training tasks. All models are pre-trained with the same pre-training GPU hours.}
\label{tab:more pre-train}
\end{table}

\paragraph{\textbf{Pre-training Tasks}} We explore the use of different pre-training tasks for masked signal modeling in Table~\ref{tab:pre-training}. Experiments reveal that MLM with a high masking ratio is sufficient for learning the interaction between vision and language. The addition of MIM further improves the results by reducing bias, as observed in~\cite{maskvlm}.  Pre-training with MIM alone results in a minimal fusion between vision and language. We hypothesize that text descriptions are typically coarse-grained and may not offer significant assistance in fine-grained vision reconstruction. Simultaneously masking both modalities and performing MIM and MLM is not recommended. This task reduces the amount of vision and language information available, which in turn increases the difficulty of MLM and MIM, resulting in performance decline. 

We further explore more pre-training tasks under the same pre-training GPU hours in Table~\ref{tab:more pre-train}. Pre-training only on MIM and MLM achieves better results in both retrieval tasks and understanding tasks, thereby demonstrating the efficiency of Masked Signal Modeling.  Performance on NLVR task is provided in Appendix.

\paragraph{\textbf{Deep FFN}} We compare different designs of FFN in the deep layers in Table~\ref{tab:routing}.
Modality-shared FFN performs better than modality-specific MoE in the deep layers, as deep features require more alignment between modalities. Using a soft router can align different modalities while obtaining more modality-specific information, thereby further enhancing performance compared to deeper architecture.
When we set $ \mathcal{L}_{aux} = 0 $, there is a noticeable decline in the model's performance across various tasks.

\begin{figure}
    \centering
    \includegraphics[width=\linewidth]{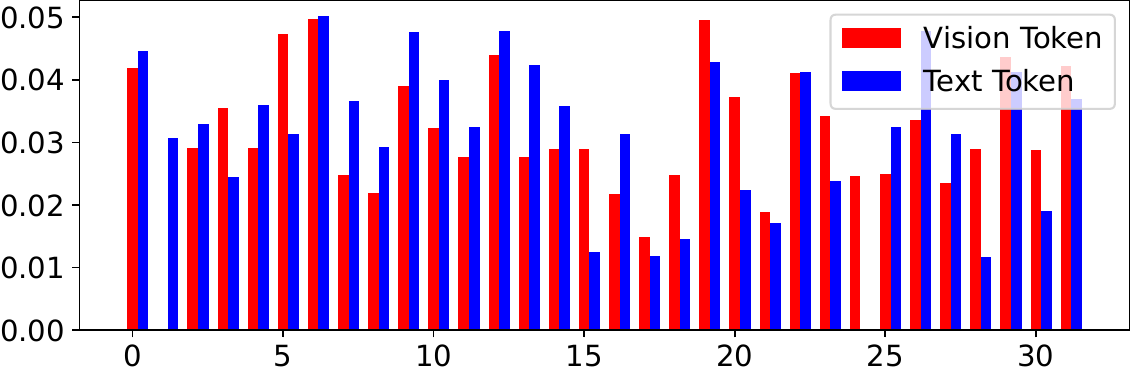}
    \caption{Frequency distribution of different modal tokens routed to specific experts in the last layer of EVE-Base.}
    \label{fig:moe_prob}
\end{figure}

Figure~\ref{fig:moe_prob} illustrates the frequency distribution of tokens from various modalities routed to specific experts in the last layer of EVE-Base during inference in retrieval tasks.
Most experts evenly process both modalities, yet some are specialized in only images or text (e.g., experts 1 and 24). 
This reveals a modality gap within the model's deeper layers. Shared FFN fails to address this issue, whereas the application of MoE can help mitigate it by routing to experts that specialize in processing vision or language tokens, thereby improving overall performance.

\begin{table}
\centering
\fontsize{9}{9}\selectfont
\begin{tabular}{l|cc|cc|c} 
\toprule
\multirow{2}{*}{Deep FFN} & \multicolumn{2}{c|}{NLVR2} & \multicolumn{2}{c|}{Flickr30K} & \multirow{2}{*}{VQA}  \\
                            & dev & test-P               & TR & IR                        \\ 
\midrule
Shared FFN & 79.6  & 80.1 & 93.5 & 80.1 & 77.0 \\
Shared FFN$^\dagger$ & 80.1  & 80.2 & 93.9 & 80.6 & 77.1 \\
Hard Router & 79.8  & 80.1 & 93.2 & 79.3 & 77.0 \\
Soft Router  & \textbf{80.3} & \textbf{80.7} & \textbf{94.4} & \textbf{81.2} & \textbf{77.4} \\
Soft Router$^\ddagger$ & 79.2  & 80.0 & 93.6 & 79.7 & 77.1 \\
\bottomrule
\end{tabular}
\caption{Ablation study on deep (top-2 layers) FFN design. Shared FFN indicates different modalities use the same FFN. We additionally add one more Transformer block to investigate the impact of parameters per token for $\dagger$. We set $\mathcal{L}_{aux}=0$ for $\ddagger$.}
\label{tab:routing}
\end{table}

\paragraph{Modality Routing} We compare the performance of the model whether use modality routing in the soft router or not in Table~\ref{tab:modality}, and the results show that our proposed modality routing can help the router to distinguish the inputs of different modalities and thus achieve better performance.

\subsection{Visualization}
We use Grad-CAM~\cite{Grad-CAM} heatmap to visualize the self-attention maps of EVE in masked signal modeling and VQA Task. Results are provided in Appendix.

\section{Conclusion}
In this paper, we present a new multimodal foundation model EVE only pre-trained by Maksed Signal Modeling with Modality-Aware MoE which is flexible and capable of encoding different modalities in a unified manner. We accelerate pre-training speed 3.5x faster than pre-training with ITC and ITM. Additionally, it is easy to scale up with a larger model or more pre-training data. Extensive experiments demonstrate that EVE outperforms existing methods in various Vision Language downstream tasks.

\begin{table}
\centering
\fontsize{9}{9}\selectfont
\begin{tabular}{l|cc|cc|c} 
\toprule
\multirow{2}{*}{Modality Routing} & \multicolumn{2}{c|}{NLVR2} & \multicolumn{2}{c|}{Flickr30K} & \multirow{2}{*}{VQA}  \\
                            & dev & test-P               & TR & IR                        \\ 
\midrule
EVE-Base & \textbf{80.3} & \textbf{80.7} & \textbf{94.4} & \textbf{81.2} & \textbf{77.4} \\
w/o MR & 79.7  & 80.0 & 93.7 & 80.8 & 77.3 \\
\bottomrule
\end{tabular}
\caption{Ablation study on modality routing technique.}
\label{tab:modality}
\end{table}

\section*{Acknowledgements}
This work was supported by the National Natural Science Foundation of China (NSFC) under Grant No. 61876224.

\bibliography{aaai24}

\begin{thebibliography}{61}
\providecommand{\natexlab}[1]{#1}

\bibitem[{Bachmann et~al.(2022)Bachmann, Mizrahi, Atanov, and Zamir}]{multimae}
Bachmann, R.; Mizrahi, D.; Atanov, A.; and Zamir, A. 2022.
\newblock MultiMAE: Multi-modal Multi-task Masked Autoencoders.
\newblock arXiv:2204.01678.

\bibitem[{Bao et~al.(2022{\natexlab{a}})Bao, Wang, Dong, Liu, Mohammed, Aggarwal, Som, Piao, and Wei}]{vlmo}
Bao, H.; Wang, W.; Dong, L.; Liu, Q.; Mohammed, O.~K.; Aggarwal, K.; Som, S.; Piao, S.; and Wei, F. 2022{\natexlab{a}}.
\newblock {VLMo}: Unified Vision-Language Pre-Training with Mixture-of-Modality-Experts.
\newblock In \emph{Advances in Neural Information Processing Systems}.

\bibitem[{Bao et~al.(2022{\natexlab{b}})Bao, Wang, Dong, and Wei}]{vlbeit}
Bao, H.; Wang, W.; Dong, L.; and Wei, F. 2022{\natexlab{b}}.
\newblock VL-BEiT: Generative Vision-Language Pretraining.
\newblock arXiv:2206.01127.

\bibitem[{Changpinyo et~al.(2021)Changpinyo, Sharma, Ding, and Soricut}]{cc12m}
Changpinyo, S.; Sharma, P.; Ding, N.; and Soricut, R. 2021.
\newblock Conceptual 12M: Pushing Web-Scale Image-Text Pre-Training To Recognize Long-Tail Visual Concepts.
\newblock In \emph{{IEEE/CVF} Conference on Computer Vision and Pattern Recognition}, 3558--3568.

\bibitem[{Chen et~al.(2020)Chen, Li, Yu, Kholy, Ahmed, Gan, Cheng, and Liu}]{uniter}
Chen, Y.; Li, L.; Yu, L.; Kholy, A.~E.; Ahmed, F.; Gan, Z.; Cheng, Y.; and Liu, J. 2020.
\newblock {UNITER:} UNiversal Image-TExt Representation Learning.
\newblock In \emph{European Conference on Computer Vision}, 104--120.

\bibitem[{Devlin et~al.(2019)Devlin, Chang, Lee, and Toutanova}]{bert}
Devlin, J.; Chang, M.; Lee, K.; and Toutanova, K. 2019.
\newblock {BERT:} Pre-training of Deep Bidirectional Transformers for Language Understanding.
\newblock In \emph{Proceedings of the 2019 Conference of the North American Chapter of the Association for Computational Linguistics}, 4171--4186.

\bibitem[{Diao et~al.(2023)Diao, Zhou, Zhang, and Wang}]{DAVINCI}
Diao, S.; Zhou, W.; Zhang, X.; and Wang, J. 2023.
\newblock Write and Paint: Generative Vision-Language Models are Unified Modal Learners.
\newblock In \emph{International Conference on Learning Representations}.

\bibitem[{Dosovitskiy et~al.(2021)Dosovitskiy, Beyer, Kolesnikov, Weissenborn, Zhai, Unterthiner, Dehghani, Minderer, Heigold, Gelly, Uszkoreit, and Houlsby}]{vit}
Dosovitskiy, A.; Beyer, L.; Kolesnikov, A.; Weissenborn, D.; Zhai, X.; Unterthiner, T.; Dehghani, M.; Minderer, M.; Heigold, G.; Gelly, S.; Uszkoreit, J.; and Houlsby, N. 2021.
\newblock An Image is Worth 16x16 Words: Transformers for Image Recognition at Scale.
\newblock In \emph{International Conference on Learning Representations}.

\bibitem[{Dou et~al.(2022{\natexlab{a}})Dou, Xu, Gan, Wang, Wang, Wang, Zhu, Zhang, Yuan, Peng, Liu, and Zeng}]{meter}
Dou, Z.; Xu, Y.; Gan, Z.; Wang, J.; Wang, S.; Wang, L.; Zhu, C.; Zhang, P.; Yuan, L.; Peng, N.; Liu, Z.; and Zeng, M. 2022{\natexlab{a}}.
\newblock An Empirical Study of Training End-to-End Vision-and-Language Transformers.
\newblock In \emph{{IEEE/CVF} Conference on Computer Vision and Pattern Recognition}, 18145--18155.

\bibitem[{Dou et~al.(2022{\natexlab{b}})Dou, Kamath, Gan, Zhang, Wang, Li, Liu, Liu, LeCun, Peng, Gao, and Wang}]{fiber}
Dou, Z.-Y.; Kamath, A.; Gan, Z.; Zhang, P.; Wang, J.; Li, L.; Liu, Z.; Liu, C.; LeCun, Y.; Peng, N.; Gao, J.; and Wang, L. 2022{\natexlab{b}}.
\newblock Coarse-to-Fine Vision-Language Pre-training with Fusion in the Backbone.
\newblock In \emph{Advances in Neural Information Processing Systems}.

\bibitem[{Duan et~al.(2022)Duan, Chen, Tran, Yang, Xu, Zeng, and Chilimbi}]{codebook}
Duan, J.; Chen, L.; Tran, S.; Yang, J.; Xu, Y.; Zeng, B.; and Chilimbi, T. 2022.
\newblock Multi-modal Alignment using Representation Codebook.
\newblock In \emph{{IEEE/CVF} Conference on Computer Vision and Pattern Recognition}, 15630--15639.

\bibitem[{Geng et~al.(2022)Geng, Liu, Lee, Schuurmans, Levine, and Abbeel}]{m3ae}
Geng, X.; Liu, H.; Lee, L.; Schuurmans, D.; Levine, S.; and Abbeel, P. 2022.
\newblock Multimodal Masked Autoencoders Learn Transferable Representations.
\newblock arXiv:2205.14204.

\bibitem[{Goyal et~al.(2017)Goyal, Khot, Summers{-}Stay, Batra, and Parikh}]{vqa}
Goyal, Y.; Khot, T.; Summers{-}Stay, D.; Batra, D.; and Parikh, D. 2017.
\newblock Making the {V} in {VQA} Matter: Elevating the Role of Image Understanding in Visual Question Answering.
\newblock In \emph{{IEEE/CVF} Conference on Computer Vision and Pattern Recognition}, 6325--6334.

\bibitem[{Gui et~al.(2023)Gui, Chang, Huang, Som, Hauptmann, Gao, and Bisk}]{vlc}
Gui, L.; Chang, Y.; Huang, Q.; Som, S.; Hauptmann, A.; Gao, J.; and Bisk, Y. 2023.
\newblock Training Vision-Language Transformers from Captions.
\newblock arXiv:2205.09256.

\bibitem[{Hazimeh et~al.(2021)Hazimeh, Zhao, Chowdhery, Sathiamoorthy, Chen, Mazumder, Hong, and Chi}]{diff}
Hazimeh, H.; Zhao, Z.; Chowdhery, A.; Sathiamoorthy, M.; Chen, Y.; Mazumder, R.; Hong, L.; and Chi, E.~H. 2021.
\newblock DSelect-k: Differentiable Selection in the Mixture of Experts with Applications to Multi-Task Learning.
\newblock In \emph{Advances in Neural Information Processing Systems}, 29335--29347.

\bibitem[{He et~al.(2022{\natexlab{a}})He, Chen, Xie, Li, Doll{\'{a}}r, and Girshick}]{mae}
He, K.; Chen, X.; Xie, S.; Li, Y.; Doll{\'{a}}r, P.; and Girshick, R.~B. 2022{\natexlab{a}}.
\newblock Masked Autoencoders Are Scalable Vision Learners.
\newblock In \emph{{IEEE/CVF} Conference on Computer Vision and Pattern Recognition}, 15979--15988.

\bibitem[{He et~al.(2022{\natexlab{b}})He, Guo, Dai, Qiao, Wu, Shu, and Ren}]{vlmae}
He, S.; Guo, T.; Dai, T.; Qiao, R.; Wu, C.; Shu, X.; and Ren, B. 2022{\natexlab{b}}.
\newblock VLMAE: Vision-Language Masked Autoencoder.
\newblock arXiv:2208.09374.

\bibitem[{Jia et~al.(2021)Jia, Yang, Xia, Chen, Parekh, Pham, Le, Sung, Li, and Duerig}]{align}
Jia, C.; Yang, Y.; Xia, Y.; Chen, Y.; Parekh, Z.; Pham, H.; Le, Q.~V.; Sung, Y.; Li, Z.; and Duerig, T. 2021.
\newblock Scaling Up Visual and Vision-Language Representation Learning With Noisy Text Supervision.
\newblock In \emph{Proceedings of the 38th International Conference on Machine Learning}, volume 139, 4904--4916.

\bibitem[{Karpathy and Fei{-}Fei(2015)}]{Karpathy}
Karpathy, A.; and Fei{-}Fei, L. 2015.
\newblock Deep visual-semantic alignments for generating image descriptions.
\newblock In \emph{{IEEE/CVF} Conference on Computer Vision and Pattern Recognition}, 3128--3137.

\bibitem[{Kim, Son, and Kim(2021)}]{vilt}
Kim, W.; Son, B.; and Kim, I. 2021.
\newblock ViLT: Vision-and-Language Transformer Without Convolution or Region Supervision.
\newblock In \emph{Proceedings of the 38th International Conference on Machine Learning}, volume 139, 5583--5594.

\bibitem[{Krishna et~al.(2017)Krishna, Zhu, Groth, Johnson, Hata, Kravitz, Chen, Kalantidis, Li, Shamma, Bernstein, and Fei{-}Fei}]{vg}
Krishna, R.; Zhu, Y.; Groth, O.; Johnson, J.; Hata, K.; Kravitz, J.; Chen, S.; Kalantidis, Y.; Li, L.; Shamma, D.~A.; Bernstein, M.~S.; and Fei{-}Fei, L. 2017.
\newblock Visual Genome: Connecting Language and Vision Using Crowdsourced Dense Image Annotations.
\newblock \emph{International Journal of Computer Vision}, 123(1): 32--73.

\bibitem[{Kwon et~al.(2023)Kwon, Cai, Ravichandran, Bas, Bhotika, and Soatto}]{maskvlm}
Kwon, G.; Cai, Z.; Ravichandran, A.; Bas, E.; Bhotika, R.; and Soatto, S. 2023.
\newblock Masked Vision and Language Modeling for Multi-modal Representation Learning.
\newblock In \emph{International Conference on Learning Representations}.

\bibitem[{Lepikhin et~al.(2021)Lepikhin, Lee, Xu, Chen, Firat, Huang, Krikun, Shazeer, and Chen}]{gshard}
Lepikhin, D.; Lee, H.; Xu, Y.; Chen, D.; Firat, O.; Huang, Y.; Krikun, M.; Shazeer, N.; and Chen, Z. 2021.
\newblock GShard: Scaling Giant Models with Conditional Computation and Automatic Sharding.
\newblock In \emph{International Conference on Learning Representations}.

\bibitem[{Lewis et~al.(2021)Lewis, Bhosale, Dettmers, Goyal, and Zettlemoyer}]{base}
Lewis, M.; Bhosale, S.; Dettmers, T.; Goyal, N.; and Zettlemoyer, L. 2021.
\newblock {BASE} Layers: Simplifying Training of Large, Sparse Models.
\newblock In \emph{Proceedings of the 38th International Conference on Machine Learning}, volume 139, 6265--6274.

\bibitem[{Li et~al.(2022)Li, Li, Xiong, and Hoi}]{blip}
Li, J.; Li, D.; Xiong, C.; and Hoi, S. 2022.
\newblock BLIP: Bootstrapping Language-Image Pre-training for Unified Vision-Language Understanding and Generation.
\newblock arXiv:2201.12086.

\bibitem[{Li et~al.(2021)Li, Selvaraju, Gotmare, Joty, Xiong, and Hoi}]{albef}
Li, J.; Selvaraju, R.~R.; Gotmare, A.; Joty, S.~R.; Xiong, C.; and Hoi, S.~C. 2021.
\newblock Align before Fuse: Vision and Language Representation Learning with Momentum Distillation.
\newblock In \emph{Advances in Neural Information Processing Systems}, 9694--9705.

\bibitem[{Li et~al.(2019)Li, Yatskar, Yin, Hsieh, and Chang}]{visualbert}
Li, L.~H.; Yatskar, M.; Yin, D.; Hsieh, C.-J.; and Chang, K.-W. 2019.
\newblock VisualBERT: A Simple and Performant Baseline for Vision and Language.
\newblock arXiv:1908.03557.

\bibitem[{Lin et~al.(2014)Lin, Maire, Belongie, Hays, Perona, Ramanan, Doll{\'{a}}r, and Zitnick}]{mscoco}
Lin, T.; Maire, M.; Belongie, S.~J.; Hays, J.; Perona, P.; Ramanan, D.; Doll{\'{a}}r, P.; and Zitnick, C.~L. 2014.
\newblock Microsoft {COCO:} Common Objects in Context.
\newblock In \emph{Proceedings of the European Conference on Computer Vision}, 740--755.

\bibitem[{Liu et~al.(2021)Liu, Zhu, Liu, Guo, Zhao, Sun, Wang, Lu, Zhou, Zhang, and Wang}]{opt}
Liu, J.; Zhu, X.; Liu, F.; Guo, L.; Zhao, Z.; Sun, M.; Wang, W.; Lu, H.; Zhou, S.; Zhang, J.; and Wang, J. 2021.
\newblock OPT: Omni-Perception Pre-Trainer for Cross-Modal Understanding and Generation.
\newblock arXiv:2107.00249.

\bibitem[{Liu et~al.(2023)Liu, Fan, Johns, Yu, Xiao, and Anandkumar}]{prismer}
Liu, S.; Fan, L.; Johns, E.; Yu, Z.; Xiao, C.; and Anandkumar, A. 2023.
\newblock Prismer: A Vision-Language Model with An Ensemble of Experts.
\newblock arXiv:2303.02506.

\bibitem[{Loshchilov and Hutter(2019)}]{adamw}
Loshchilov, I.; and Hutter, F. 2019.
\newblock Decoupled Weight Decay Regularization.
\newblock In \emph{International Conference on Learning Representations}.

\bibitem[{Mustafa et~al.(2022)Mustafa, Riquelme, Puigcerver, Jenatton, and Houlsby}]{LIMoE}
Mustafa, B.; Riquelme, C.; Puigcerver, J.; Jenatton, R.; and Houlsby, N. 2022.
\newblock Multimodal Contrastive Learning with LIMoE: the Language-Image Mixture of Experts.
\newblock In \emph{Advances in Neural Information Processing Systems}.

\bibitem[{Ordonez, Kulkarni, and Berg(2011)}]{sbu}
Ordonez, V.; Kulkarni, G.; and Berg, T.~L. 2011.
\newblock Im2Text: Describing Images Using 1 Million Captioned Photographs.
\newblock In \emph{Advances in Neural Information Processing Systems}, 1143--1151.

\bibitem[{Ouyang et~al.(2022)Ouyang, Wu, Jiang, Almeida, Wainwright, Mishkin, Zhang, Agarwal, Slama, Ray, Schulman, Hilton, Kelton, Miller, Simens, Askell, Welinder, Christiano, Leike, and Lowe}]{instructgpt}
Ouyang, L.; Wu, J.; Jiang, X.; Almeida, D.; Wainwright, C.~L.; Mishkin, P.; Zhang, C.; Agarwal, S.; Slama, K.; Ray, A.; Schulman, J.; Hilton, J.; Kelton, F.; Miller, L.; Simens, M.; Askell, A.; Welinder, P.; Christiano, P.; Leike, J.; and Lowe, R. 2022.
\newblock Training language models to follow instructions with human feedback.
\newblock arXiv:2203.02155.

\bibitem[{Peng et~al.(2022)Peng, Dong, Bao, Ye, and Wei}]{beitv2}
Peng, Z.; Dong, L.; Bao, H.; Ye, Q.; and Wei, F. 2022.
\newblock BEiT v2: Masked Image Modeling with Vector-Quantized Visual Tokenizers.
\newblock arXiv:2208.06366.

\bibitem[{Plummer et~al.(2015)Plummer, Wang, Cervantes, Caicedo, Hockenmaier, and Lazebnik}]{flickr}
Plummer, B.~A.; Wang, L.; Cervantes, C.~M.; Caicedo, J.~C.; Hockenmaier, J.; and Lazebnik, S. 2015.
\newblock Flickr30k Entities: Collecting Region-to-Phrase Correspondences for Richer Image-to-Sentence Models.
\newblock In \emph{{IEEE/CVF} International Conference on Computer Vision}, 2641--2649.

\bibitem[{Radford et~al.(2021)Radford, Kim, Hallacy, Ramesh, Goh, Agarwal, Sastry, Askell, Mishkin, Clark, Krueger, and Sutskever}]{clip}
Radford, A.; Kim, J.~W.; Hallacy, C.; Ramesh, A.; Goh, G.; Agarwal, S.; Sastry, G.; Askell, A.; Mishkin, P.; Clark, J.; Krueger, G.; and Sutskever, I. 2021.
\newblock Learning Transferable Visual Models From Natural Language Supervision.
\newblock In \emph{Proceedings of the 38th International Conference on Machine Learning}, volume 139, 8748--8763.

\bibitem[{Ramesh et~al.(2021)Ramesh, Pavlov, Goh, Gray, Voss, Radford, Chen, and Sutskever}]{dalle}
Ramesh, A.; Pavlov, M.; Goh, G.; Gray, S.; Voss, C.; Radford, A.; Chen, M.; and Sutskever, I. 2021.
\newblock Zero-Shot Text-to-Image Generation.
\newblock In Meila, M.; and Zhang, T., eds., \emph{Proceedings of the 38th International Conference on Machine Learning}, volume 139, 8821--8831.

\bibitem[{Riquelme et~al.(2021)Riquelme, Puigcerver, Mustafa, Neumann, Jenatton, Pinto, Keysers, and Houlsby}]{vmoe}
Riquelme, C.; Puigcerver, J.; Mustafa, B.; Neumann, M.; Jenatton, R.; Pinto, A.~S.; Keysers, D.; and Houlsby, N. 2021.
\newblock Scaling Vision with Sparse Mixture of Experts.
\newblock In \emph{Advances in Neural Information Processing Systems}, 8583--8595.

\bibitem[{Selvaraju et~al.(2017)Selvaraju, Cogswell, Das, Vedantam, Parikh, and Batra}]{Grad-CAM}
Selvaraju, R.~R.; Cogswell, M.; Das, A.; Vedantam, R.; Parikh, D.; and Batra, D. 2017.
\newblock Grad-CAM: Visual Explanations from Deep Networks via Gradient-Based Localization.
\newblock In \emph{{IEEE/CVF} International Conference on Computer Vision}, 618--626.

\bibitem[{Sharma et~al.(2018)Sharma, Ding, Goodman, and Soricut}]{cc}
Sharma, P.; Ding, N.; Goodman, S.; and Soricut, R. 2018.
\newblock Conceptual Captions: {A} Cleaned, Hypernymed, Image Alt-text Dataset For Automatic Image Captioning.
\newblock In \emph{Proceedings of the 56th Annual Meeting of the Association for Computational Linguistics}, 2556--2565.

\bibitem[{Shazeer et~al.(2017)Shazeer, Mirhoseini, Maziarz, Davis, Le, Hinton, and Dean}]{auxloss}
Shazeer, N.; Mirhoseini, A.; Maziarz, K.; Davis, A.; Le, Q.~V.; Hinton, G.~E.; and Dean, J. 2017.
\newblock Outrageously Large Neural Networks: The Sparsely-Gated Mixture-of-Experts Layer.
\newblock In \emph{International Conference on Learning Representations}.

\bibitem[{Shen et~al.(2023)Shen, Yao, Li, Darrell, Keutzer, and He}]{vlmoe}
Shen, S.; Yao, Z.; Li, C.; Darrell, T.; Keutzer, K.; and He, Y. 2023.
\newblock Scaling Vision-Language Models with Sparse Mixture of Experts.
\newblock In \emph{The 2023 Conference on Empirical Methods in Natural Language Processing}.

\bibitem[{Su et~al.(2020)Su, Zhu, Cao, Li, Lu, Wei, and Dai}]{vlbert}
Su, W.; Zhu, X.; Cao, Y.; Li, B.; Lu, L.; Wei, F.; and Dai, J. 2020.
\newblock {VL-BERT:} Pre-training of Generic Visual-Linguistic Representations.
\newblock In \emph{International Conference on Learning Representations}.

\bibitem[{Suhr et~al.(2019)Suhr, Zhou, Zhang, Zhang, Bai, and Artzi}]{nlvr}
Suhr, A.; Zhou, S.; Zhang, A.; Zhang, I.; Bai, H.; and Artzi, Y. 2019.
\newblock A Corpus for Reasoning about Natural Language Grounded in Photographs.
\newblock In \emph{Proceedings of the 57th Conference of the Association for Computational Linguistics}, 6418--6428.

\bibitem[{Vaswani et~al.(2017)Vaswani, Shazeer, Parmar, Uszkoreit, Jones, Gomez, Kaiser, and Polosukhin}]{trans}
Vaswani, A.; Shazeer, N.; Parmar, N.; Uszkoreit, J.; Jones, L.; Gomez, A.~N.; Kaiser, L.; and Polosukhin, I. 2017.
\newblock Attention is All you Need.
\newblock In \emph{Advances in Neural Information Processing Systems}, 5998--6008.

\bibitem[{Wang et~al.(2022{\natexlab{a}})Wang, Yang, Hu, Li, Lin, Gan, Liu, Liu, and Wang}]{git}
Wang, J.; Yang, Z.; Hu, X.; Li, L.; Lin, K.; Gan, Z.; Liu, Z.; Liu, C.; and Wang, L. 2022{\natexlab{a}}.
\newblock {GIT}: A Generative Image-to-text Transformer for Vision and Language.
\newblock \emph{Transactions on Machine Learning Research}.

\bibitem[{Wang et~al.(2023)Wang, Bao, Dong, Bjorck, Peng, Liu, Aggarwal, Mohammed, Singhal, Som, and Wei}]{beit3}
Wang, W.; Bao, H.; Dong, L.; Bjorck, J.; Peng, Z.; Liu, Q.; Aggarwal, K.; Mohammed, O.~K.; Singhal, S.; Som, S.; and Wei, F. 2023.
\newblock Image as a foreign language: {BEiT} pretraining for vision and vision-language tasks.
\newblock In \emph{{IEEE/CVF} Conference on Computer Vision and Pattern Recognition}.

\bibitem[{Wang et~al.(2022{\natexlab{b}})Wang, Yu, Yu, Dai, Tsvetkov, and Cao}]{simvlm}
Wang, Z.; Yu, J.; Yu, A.~W.; Dai, Z.; Tsvetkov, Y.; and Cao, Y. 2022{\natexlab{b}}.
\newblock SimVLM: Simple Visual Language Model Pretraining with Weak Supervision.
\newblock In \emph{International Conference on Learning Representations}.

\bibitem[{Wei et~al.(2022{\natexlab{a}})Wei, Fan, Xie, Wu, Yuille, and Feichtenhofer}]{maskfeat}
Wei, C.; Fan, H.; Xie, S.; Wu, C.; Yuille, A.~L.; and Feichtenhofer, C. 2022{\natexlab{a}}.
\newblock Masked Feature Prediction for Self-Supervised Visual Pre-Training.
\newblock In \emph{{IEEE/CVF} Conference on Computer Vision and Pattern Recognition}, 14648--14658.

\bibitem[{Wei et~al.(2022{\natexlab{b}})Wei, Tay, Bommasani, Raffel, Zoph, Borgeaud, Yogatama, Bosma, Zhou, Metzler, Chi, Hashimoto, Vinyals, Liang, Dean, and Fedus}]{emerge}
Wei, J.; Tay, Y.; Bommasani, R.; Raffel, C.; Zoph, B.; Borgeaud, S.; Yogatama, D.; Bosma, M.; Zhou, D.; Metzler, D.; Chi, E.~H.; Hashimoto, T.; Vinyals, O.; Liang, P.; Dean, J.; and Fedus, W. 2022{\natexlab{b}}.
\newblock Emergent Abilities of Large Language Models.
\newblock arXiv:2206.07682.

\bibitem[{Yang et~al.(2022)Yang, Duan, Tran, Xu, Chanda, Chen, Zeng, Chilimbi, and Huang}]{triple}
Yang, J.; Duan, J.; Tran, S.; Xu, Y.; Chanda, S.; Chen, L.; Zeng, B.; Chilimbi, T.; and Huang, J. 2022.
\newblock Vision-Language Pre-Training with Triple Contrastive Learning.
\newblock In \emph{{IEEE/CVF} Conference on Computer Vision and Pattern Recognition}, 15650--15659.

\bibitem[{Yao et~al.(2022)Yao, Huang, Hou, Lu, Niu, Xu, Liang, Li, Jiang, and Xu}]{filip}
Yao, L.; Huang, R.; Hou, L.; Lu, G.; Niu, M.; Xu, H.; Liang, X.; Li, Z.; Jiang, X.; and Xu, C. 2022.
\newblock {FILIP}: Fine-grained Interactive Language-Image Pre-Training.
\newblock In \emph{International Conference on Learning Representations}.

\bibitem[{Yuan et~al.(2021)Yuan, Chen, Chen, Codella, Dai, Gao, Hu, Huang, Li, Li, Liu, Liu, Liu, Lu, Shi, Wang, Wang, Xiao, Xiao, Yang, Zeng, Zhou, and Zhang}]{florence}
Yuan, L.; Chen, D.; Chen, Y.-L.; Codella, N.; Dai, X.; Gao, J.; Hu, H.; Huang, X.; Li, B.; Li, C.; Liu, C.; Liu, M.; Liu, Z.; Lu, Y.; Shi, Y.; Wang, L.; Wang, J.; Xiao, B.; Xiao, Z.; Yang, J.; Zeng, M.; Zhou, L.; and Zhang, P. 2021.
\newblock Florence: A New Foundation Model for Computer Vision.
\newblock arXiv:2111.11432.

\bibitem[{Zeng, Zhang, and Li(2022)}]{xvlm}
Zeng, Y.; Zhang, X.; and Li, H. 2022.
\newblock Multi-Grained Vision Language Pre-Training: Aligning Texts with Visual Concepts.
\newblock In \emph{Proceedings of the 39th International Conference on Machine Learning}, volume 162, 25994--26009.

\bibitem[{Zhang et~al.(2021)Zhang, Li, Hu, Yang, Zhang, Wang, Choi, and Gao}]{vinvl}
Zhang, P.; Li, X.; Hu, X.; Yang, J.; Zhang, L.; Wang, L.; Choi, Y.; and Gao, J. 2021.
\newblock VinVL: Revisiting Visual Representations in Vision-Language Models.
\newblock In \emph{{IEEE/CVF} Conference on Computer Vision and Pattern Recognition}, 5579--5588.

\bibitem[{Zhang et~al.(2023)Zhang, Zeng, Zhang, and Li}]{xfm}
Zhang, X.; Zeng, Y.; Zhang, J.; and Li, H. 2023.
\newblock Toward Building General Foundation Models for Language, Vision, and Vision-Language Understanding Tasks.
\newblock arXiv:2301.05065.

\bibitem[{Zhao et~al.(2023{\natexlab{a}})Zhao, Guo, He, Shao, Yuan, and Liu}]{mamo}
Zhao, Z.; Guo, L.; He, X.; Shao, S.; Yuan, Z.; and Liu, J. 2023{\natexlab{a}}.
\newblock MAMO: Fine-Grained Vision-Language Representations Learning with Masked Multimodal Modeling.
\newblock In \emph{Proceedings of the 46th International ACM SIGIR Conference on Research and Development in Information Retrieval}.

\bibitem[{Zhao et~al.(2023{\natexlab{b}})Zhao, Guo, Yue, Chen, Shao, Zhu, Yuan, and Liu}]{chatbridge}
Zhao, Z.; Guo, L.; Yue, T.; Chen, S.; Shao, S.; Zhu, X.; Yuan, Z.; and Liu, J. 2023{\natexlab{b}}.
\newblock ChatBridge: Bridging Modalities with Large Language Model as a Language Catalyst.
\newblock arXiv:2305.16103.

\bibitem[{Zhu et~al.(2022)Zhu, Zhu, Wang, Wang, Li, Wang, and Dai}]{uni-moe}
Zhu, J.; Zhu, X.; Wang, W.; Wang, X.; Li, H.; Wang, X.; and Dai, J. 2022.
\newblock Uni-Perceiver-MoE: Learning Sparse Generalist Models with Conditional MoEs.
\newblock In \emph{Advances in Neural Information Processing Systems}.

\bibitem[{Zoph et~al.(2022)Zoph, Bello, Kumar, Du, Huang, Dean, Shazeer, and Fedus}]{zloss}
Zoph, B.; Bello, I.; Kumar, S.; Du, N.; Huang, Y.; Dean, J.; Shazeer, N.; and Fedus, W. 2022.
\newblock ST-MoE: Designing Stable and Transferable Sparse Expert Models.
\newblock arXiv:2202.08906.

\end{thebibliography}

\clearpage
\appendix

\begin{table}
\centering
\fontsize{9}{9}\selectfont
\begin{tabular}{l|c|c|c}
\toprule
Method  & \makecell[c]{Inference\\Time} & \makecell[c]{\#Params\\Per Token} & \makecell[c]{VQA} \\ \midrule
ALBEF \cite{albef}             & 35.3 min  & 210M & 74.54     \\
VLMO \cite{vlmo}               & 21.6 min  & 180M & 76.64 \\
EVE-Base (Ours)                & 24.8 min  & 190M & \textbf{78.00}  \\ 
\bottomrule
\end{tabular}
\caption{VQA test set inference time, parameters per token, and VQA test-dev accuracy of different methods on 8 V100 GPUs. The inference time of other methods is reproduced by us.}
\label{tab:inftime}
\end{table}

\begin{table}
\centering
\begin{tabular}{cccc|cc}
\toprule
\multicolumn{4}{c|}{Pre-training Tasks} & \multicolumn{2}{c}{NLVR2} \\
MIM & MLM & ITC & ITM & dev & test-P \\ \midrule
\usym{2713} & \usym{2713} & \usym{2713} & & 79.5 & 79.4 \\
\usym{2713} & \usym{2713} & & \usym{2713} & 81.4 & 81.7 \\
\usym{2713} & \usym{2713} & \usym{2713} & \usym{2713} & \textbf{81.6} & 81.8 \\
\usym{2713} & \usym{2713} & & & \textbf{81.6} & \textbf{82.8} \\
\bottomrule 
\end{tabular}
\caption{Ablation study on more pre-training tasks. All models are pre-trained with the same pre-training GPU hours. We use the model fine-tuned on retrieval task for initialization.}
\label{tab:more pre-train nlvr}
\end{table}

\begin{table}[!htbp]
\centering
\begin{tabular}{c|cc|cc|c} 
\toprule
\multirow{2}{*}{\makecell[c]{Decoder\\Depth}} & \multicolumn{2}{c|}{NLVR2} & \multicolumn{2}{c|}{Flickr30K} & \multirow{2}{*}{VQA}  \\
                            & dev & test-P                                             & TR & IR                        \\ \midrule
2    & \textbf{79.7}  & 80.2 & 93.6 & 80.1 & 77.2 \\
4    & 79.3  & \textbf{80.4} & 93.5 & 80.4 & 77.1 \\
8    & \textbf{79.7}  & 80.1 & \textbf{93.9} & \textbf{80.7} & \textbf{77.3} \\
12   & 79.2  & 80.1 & 93.0 & 79.8 & 77.1 \\
\bottomrule
\end{tabular}
\caption{Ablation study on MIM decoder depth.}
\label{tab:decdepth}
\end{table}

\begin{table}
\centering
\begin{tabular}{l|cc|cc|c} 
\toprule
\multirow{2}{*}{Position} & \multicolumn{2}{c|}{NLVR2} & \multicolumn{2}{c|}{Flickr30K} & \multirow{2}{*}{VQA}  \\
                            & dev & test-P                                             & TR & IR                        \\ 
\midrule
{[}11,12]          & \textbf{79.7}  & \textbf{80.1} & \textbf{93.9} & \textbf{80.7} & \textbf{77.3} \\
{[}10,12]          & 79.6  & 79.8 & 93.3 & 80.4 & 77.2 \\
{[}6,7]            & 78.6  & 79.2 & 92.3 & 78.4 & 76.7 \\
{[}1,2]            & 79.0  & 79.7 & 92.8 & 78.7 & 76.9 \\
\bottomrule
\end{tabular}
\caption{Ablation study on the position of soft router. Transformer blocks with soft router are shown in list form.}
\label{tab:moelayer}
\end{table}

\begin{table}
\centering
\begin{tabular}{l|cc|cc|c} 
\toprule
\multirow{2}{*}{Shallow FFN} & \multicolumn{2}{c|}{NLVR2} & \multicolumn{2}{c|}{Flickr30K} & \multirow{2}{*}{VQA}  \\
                            & dev & test-P               & TR & IR                        \\ 
\midrule
Shared FFN              & 79.3  & 79.6 & 93.7 & 80.5 & 76.6      \\
Hard Router             & 79.7  & 80.1 & 93.9 & 80.7 & 77.3      \\
Soft Router   & \textbf{80.8}  & \textbf{80.5} & \textbf{95.0} & \textbf{81.1} & \textbf{77.5}      \\
\bottomrule
\end{tabular}
\caption{Ablation study on shallow (bottom-10 layers) FFN design. Shared FFN indicates different modalities use the same FFN.}
\label{tab:ffn}
\end{table}

\section{Pre-training and Inference Speed}
We present pre-training time in Figure~\ref{fig:gputime}, while inference time and parameters per token are in Table in Appendix. We exclude the data preprocessing time in inference for a fair comparison. Our method surpasses other methods in pre-training speed, especially compared to VL-BEiT~\cite{vlbeit} and VLMO~\cite{vlmo} by a large margin.
We test the inference speed on VQA test set.
After adding the MoE, EVE's inference time does not increase significantly and achieves the best performance, only slightly slower than VLMO but much faster than ALBEF. This phenomenon is consistent with the parameters per token of these models.

\section{More Ablation Studies}

\paragraph{\textbf{MIM Decoder Depth}} In Table~\ref{tab:decdepth}, we compare the different depths of the decoder in masked image modeling. Due to full supervision of downstream tasks, various decoder designs have no noticeable difference, but the performance drops slightly with a too-deep decoder.

\paragraph{\textbf{Position of Soft Router}} We also explore the impact of the position of soft router in Table~\ref{tab:moelayer}. The experimental results indicate that the performance is higher when the soft router is placed in the top layers, followed by the bottom layers, and then the middle layers. Using a soft router across layers yields slightly lower performance than continuous use.  We argue that high-level features are relatively uniform, while the modality information of low-level features is relatively more pronounced, making it easier for the router to process. Additionally, high-level features require more fusion, resulting in better performance when the soft router is placed in the top layers.

\paragraph{\textbf{Shallow FFN}} 
Table~\ref{tab:ffn} presents the results of different designs of FFN in the shallow layers. Experimental results show that using modality-specific MoE to obtain modality information in shallow layers can achieve better results than using modality-shared FFN, which emphasizes more on modality fusion. Using a soft router can combine the advantages of both approaches and exhibit promising performance, but this will lead to an increase in computational overhead.

\section{Model Architecture Implementation Details}
EVE-Base consists of 12 Transformer blocks with 12 attention heads and 768 hidden size. EVE-Large consists of 24 Transformer blocks with 16 attention heads and 1024 hidden size. We employ a soft router with 32 experts in EVE-Base on top-2 layers, EVE-Large on top-3 layers, and a hard router on the other layers. We pre-train EVE-Base for 480k steps with a batch size of 2048 on 16 A100 GPUs and EVE-Large with the same batch size for 280k steps on 32 A100 GPUs. We use AdamW~\cite{adamw} optimizer with $\beta_1 = 0.9$, $\beta_2 = 0.98$ and a weight decay of 0.05. The peak learning rate is 5e-4 for EVE-Base and 2e-4 for EVE-Large. We use linear warmup for the first 10k steps and cosine learning rate decay. During pre-training, the image resolution is $224\times 224$, and the patch size is $16\times 16$. We use random resized cropping and  horizontal flipping for data augmentation. We mask 75\% of image patches by random sampling in masked image modeling and 50\% of text in masked language modeling. EVE is initialized with BEiTv2 \cite{beitv2}. Mixed-precision is used for pre-training.

\section{Downstream Tasks Fine-tuning Details}
We use the model fine-tuned on MSCOCO Retrieval Task for other downstream tasks. We use AdamW~\cite{adamw} optimizer with a weight decay of 0.01 and a cosine learning rate scheduler. We use linear warmup in the first 10\% steps during fine-tuning. The input image resolution is $480\times 480$ for VQA and $384\times 384$ for other tasks.

\subsection{Image-Text Understanding Task}
\paragraph{\textbf{Visual Question Answering (VQA)}}
Following previous methods~\cite{albef, mamo}, we use both train and validation sets in VQA2.0 dataset~\cite{vqa} for training, and we do not use question-answer pairs from Visual Genome~\cite{vg} for augmentation. Following~\citet{vlmo, beit3}, we view the task as a classification task to choose the answer from a set of size 3129. $\boldsymbol{T}_{\text{cls}}$ token is used as the representation of the image-question pair and fed into a two-layer classifier to predict the answer.  
We fine-tune the models for 10 epochs with 128 batch size and a peak learning rate of 3e-5.

\paragraph{\textbf{Natural Language for Visual Reasoning (NLVR2)}}
Following~\citet{albef, mamo}, we convert the triplet input into two image-text pairs with the same text description and different images and extract their multimodal representation separately. We use a bi-attention module to fuse two multimodal representations following~\citet{mamo}. We concatenate fused representations and use a two-layer classifier to predict the answer.
We fine-tune the models for 10 epochs with 128 batch size. The peak learning rate is 4e-5 for EVE-Base and 3e-5 for EVE-Large.

\subsection{Image-Text Retrieval Task}
We split MSCOCO and Flickr30K into train, validation, and test sets following widely used Karpathy split~\cite{Karpathy}.
During inference, we select the top-128 candidates using ITC scores and rerank them based on ITM score following~\citet{albef,mamo}.
\paragraph{\textbf{MSCOCO Image-Text Retrieval}}
We fine-tune the models with 256 batch size for 10 epochs. The peak learning rate is 3.5e-5 for the base model and 5e-5 for the large model.

\paragraph{\textbf{Flickr Image-Text Retrieval}}
We fine-tune the models with 128 batch size for 10 epochs and a peak learning rate of 1.5e-5.

\section{Visualization}
We use Grad-CAM~\cite{Grad-CAM} heatmap to visualize the self-attention maps of EVE in masked signal modeling and VQA Task. We employ Grad-CAM visualization for both masked image patches and masked text tokens on the last layer of EVE.
For MLM, we mask a keyword and take the Grad-CAM heatmap of the corresponding image as the result. For MIM, we randomly mask 75\% patches as pre-training and take the Grad-CAM values of corresponding text as the result.
It is obvious that EVE pays more attention to information from other modalities that is semantically related to the masked portion. This reflects that complementary information can be learned between different modalities through simple masked signal modeling, without necessarily having to use complex pre-training tasks.
\subsection{Visualization on Masked Language Modeling}
\begin{figure*}
    \vspace{2cm}
    \centering
    \includegraphics[width=0.95\linewidth]{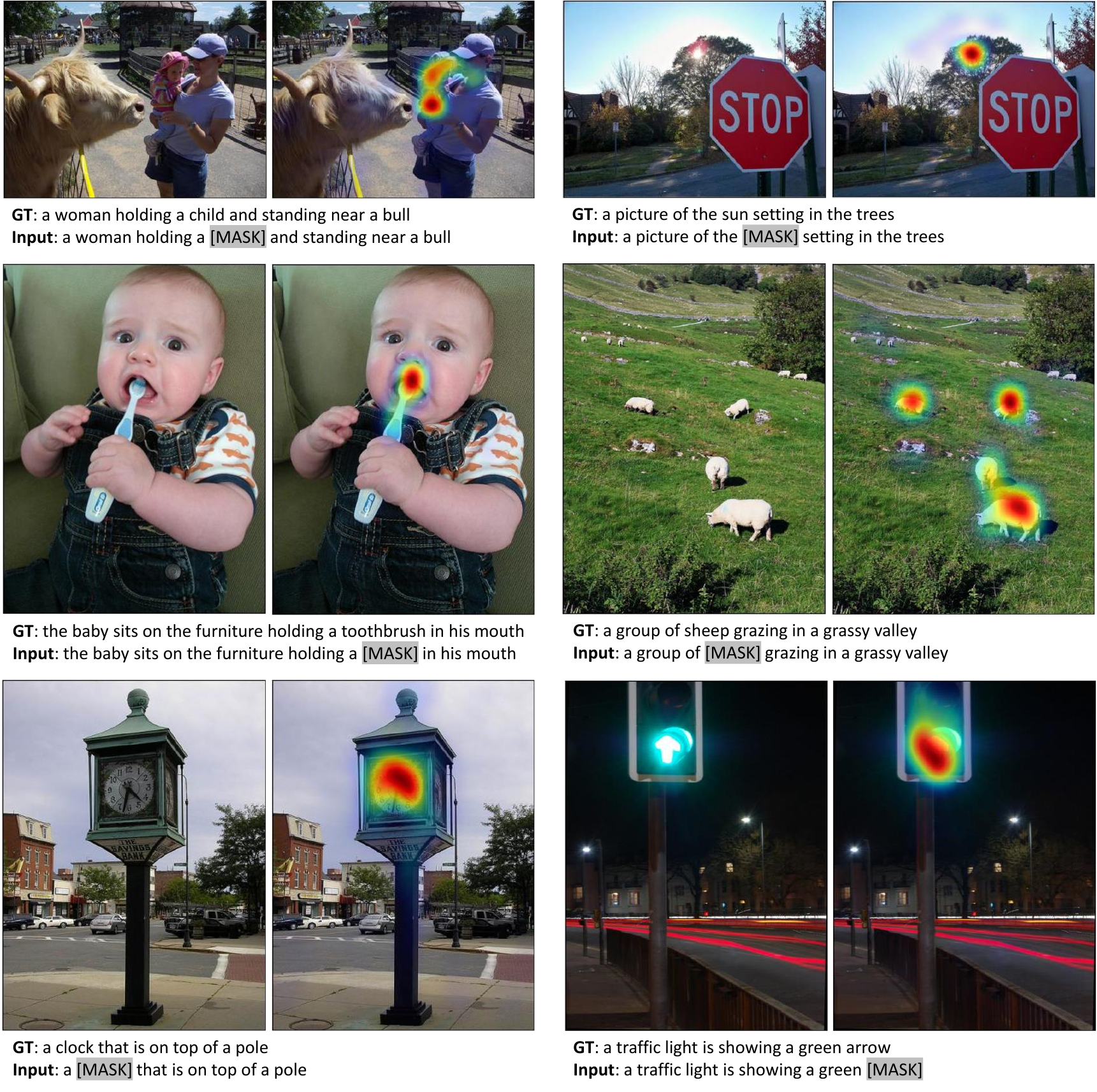}
    \caption{More examples of Grad-CAM visualization of Masked Language Modeling on masked text tokens. EVE pays more attention to the masked word in the image to reconstruct the word.}
    \label{fig:mlm_appendix}
\end{figure*}

In Figure~\ref{fig:mlm_appendix} we present some examples of Grad-CAM~\cite{Grad-CAM} visualization on a masked keyword in Masked Language Modeling. The heatmaps on images are from the last layer of EVE-Base.

\subsection{Visualization on Masked Image Modeling}
\begin{figure*}
    \vspace{1.5cm}
    \centering
    \includegraphics[width=0.95\linewidth]{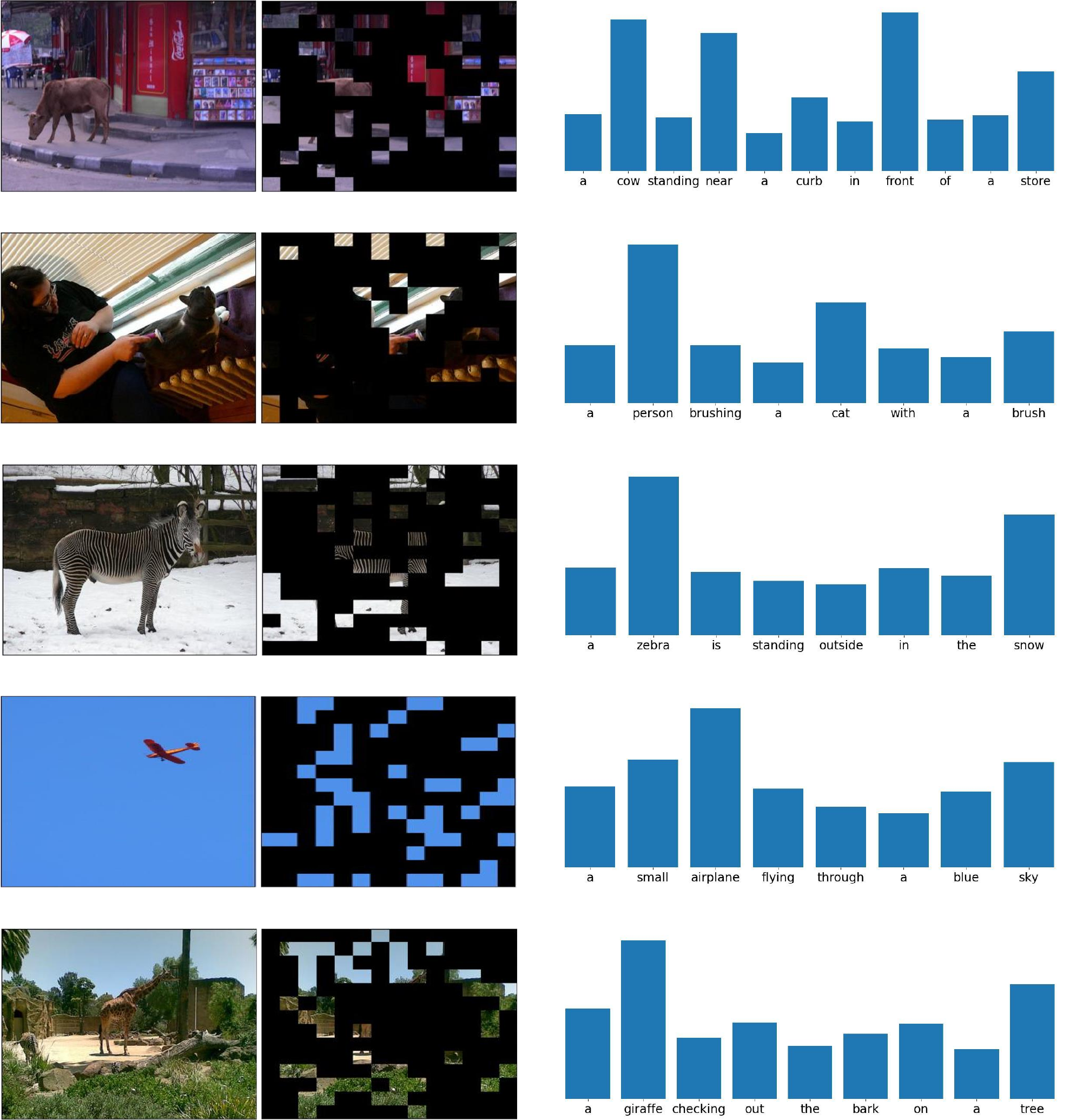}
   \caption{More examples of Grad-CAM visualization of Masked Image Modeling on masked image patches. We present the Grad-CAM value of each word in the histogram. Similar to MLM, EVE places more emphasis on the masked image regions described by the text.}
    \label{fig:mim_appendix}
\end{figure*}

In Figure~\ref{fig:mim_appendix} we present more examples of Grad-CAM~\cite{Grad-CAM} visualization on masked image patches in Masked Image Modeling. The weights on texts are from the last layer of EVE-Base.

\subsection{Visualization on VQA}
\begin{figure*}
    \vspace{2cm}
    \centering
    \includegraphics[width=0.95\linewidth]{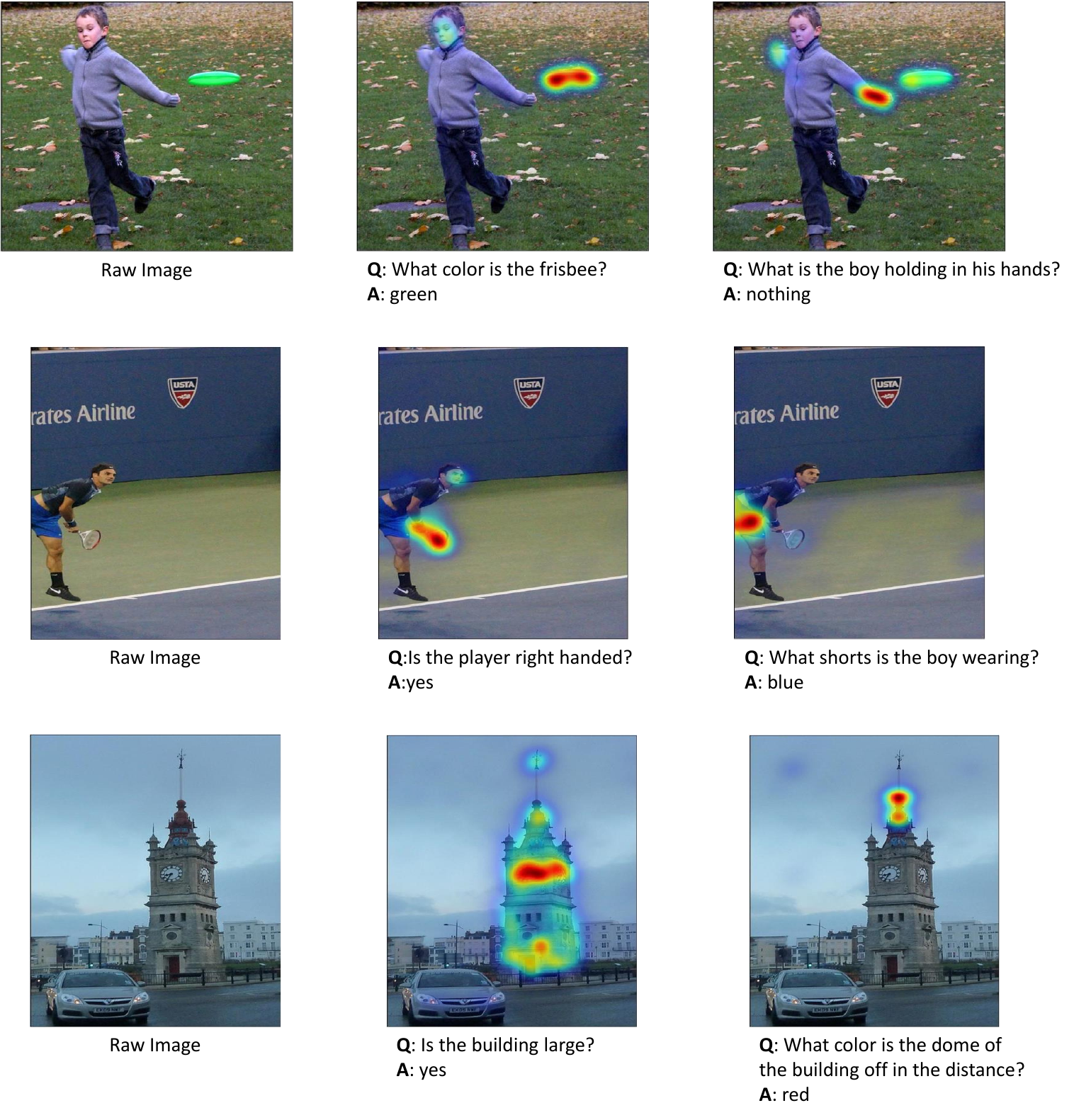}
    \caption{Grad-CAM visualization on VQA. It is clear that EVE focuses on the critical regions in the image that can answer the question.}
    \label{fig:vqa_appendix}
\end{figure*}

We show some examples of the Grad-CAM~\cite{Grad-CAM} heatmap from the last layer of EVE-Base on VQA in Figure~\ref{fig:vqa_appendix}.

\end{document}